\documentclass[final,journal,letterpaper,twocolumn]{IEEEtran}

\ifCLASSINFOpdf
  \usepackage[pdftex]{graphicx}
  \graphicspath{{./figures/}}
  \DeclareGraphicsExtensions{.eps,.pdf,.jpeg,.png,.tif}
\else
\fi

\usepackage{amsmath}

\usepackage{algorithm,algorithmic}

\usepackage{booktabs}

\usepackage{xcolor}

\usepackage{array}

\ifCLASSOPTIONcompsoc
  \usepackage[caption=false,font=normalsize,labelfont=sf,textfont=sf]{subfig}
\else
  \usepackage[caption=false,font=footnotesize]{subfig}
\fi

\begin{document}

\title{Building Footprint Generation by Integrating Convolution Neural Network with Feature Pairwise Conditional Random Field (FPCRF)}


\author{Qingyu Li,
        Yilei Shi,~\IEEEmembership{Member,~IEEE},
        Xin Huang,~\IEEEmembership{Senior Member,~IEEE}, \newline
        and~Xiao Xiang~Zhu,~\IEEEmembership{Senior~Member,~IEEE}
\thanks{This work is supported by the European Research Council (ERC) under the European Union's Horizon 2020 research and innovation programme (grant agreement no. ERC-2016-StG-714087, acronym: So2Sat, www.so2sat.eu), the Helmholtz Association under the framework of the Young Investigators Group ``SiPEO" (VH-NG-1018, www.sipeo.bgu.tum.de) and Helmholtz Excellent Professorship ``Data Science in Earth Observation - Big Data Fusion for Urban Research''. The authors thank the Gauss Centre for Supercomputing (GCS) e.V. for funding this project by providing computing time on the GCS Supercomputer SuperMUC at the Leibniz Supercomputing Centre (LRZ) and on the supercomputer JURECA at Forschungszentrum J{\"u}lich. The authors thank Planet and the Defence Science and Technology Laboratory (DSTL) for providing the datasets.}

\thanks{Q.~Li is with the Remote Sensing Technology Institute (IMF), German Aerospace Center (DLR), 82234 We\ss ling, Germany and Signal Processing in Earth Observation, Technische Universit{\"a}t M{\"u}nchen (TUM), 80333 Munich, Germany (e-mail: qingyu.li@dlr.de)}

\thanks{Y.~Shi is with the Chair of Remote Sensing Technology, Technische Universit{\"a}t M{\"u}nchen (TUM), 80333 Munich, Germany (e-mail: yilei.shi@tum.de)}

\thanks{X.~Huang is with the Department of Remote Sensing, School of Remote Sensing and Information Engineering, Wuhan University, Wuhan, China (e-mail: xhuang@whu.edu.cn)}
\thanks{X.~Zhu is with the Remote Sensing Technology Institute (IMF), German Aerospace Center (DLR), 82234 We\ss ling, Germany
 and Signal Processing in Earth Observation, Technische Universit{\"a}t M{\"u}nchen (TUM), 80333 Munich, Germany (e-mail: xiaoxiang.zhu@dlr.de)}
 \thanks{\emph{(Correspondence: Xiao Xiang Zhu)}}
}

%
%

\markboth{submitted to IEEE TRANSACTIONS ON GEOSCIENCE AND REMOTE SENSING, 2019}
{A. B \MakeLowercase{\textit{et al.}}: Building footprint generation by integrating convolution neural network with feature pairwise conditional random field (FPCRF)}
%

\maketitle

\begin{abstract}
\textcolor{blue}{This is the preprint version, to read the final version please go to IEEE Transactions on Geoscience and Remote Sensing  on  IEEE  Xplore.} Building footprint maps are vital to many remote sensing applications, such as 3D building modeling, urban planning, and disaster management. Due to the complexity of buildings, the accurate and reliable generation of the building footprint from remote sensing imagery is still a challenging task. In this work, an end-to-end building footprint generation approach that integrates convolution neural network (CNN) and graph model is proposed. CNN serves as the feature extractor, while the graph model can take spatial correlation into consideration. Moreover, we propose to implement the feature pairwise conditional random field (FPCRF) as a graph model to preserve sharp boundaries and fine-grained segmentation. Experiments are conducted on four different datasets: (1) Planetscope satellite imagery of the cities of Munich, Paris, Rome, and Zurich; (2) ISPRS benchmark data from the city of Potsdam, (3) Dstl Kaggle dataset; and (4) Inria Aerial Image Labeling data of Austin, Chicago, Kitsap County, Western Tyrol, and Vienna. It is found that the proposed end-to-end building footprint generation framework with the FPCRF as the graph model can further improve the accuracy of building footprint generation by using only CNN, which is the current state-of-the-art.
\end{abstract}


\section{Introduction}
Building footprint generation is an active field of research with the domain of remote sensing (RS). The established building footprint maps are useful to understand urban dynamics in many important applications, and also facilitate the assessment of the extent of damages after natural disasters such as earthquakes. OpenStreetMap (OSM) can provide manually annotated building footprint information for some urban areas, however, it is not  always available in many parts of the world.  Therefore, high-resolution RS imagery, which covers global areas and contains huge potential for meaningful ground information extraction, is a reliable source of data for building footprint generation. However, automatic building footprint generation from high-resolution RS imagery is still difficult because of variations in the appearance of buildings, complicated background interference, shooting angle, shadows, and illumination conditions. Moreover, buildings and the other impervious objects in urban areas have similar spectral and spatial characteristics. 

Early studies of automatic building footprint generation from high resolution RS imagery rely on regular shape and line segments of buildings to recognize buildings. Line segments of the building are first detected and extracted by edge drawing lines (EDLines) \cite{akinlar2011edlines}, and then hierarchically grouped into candidate rectangular buildings by a graph search-based perceptual grouping approach in \cite{wang2015efficient}. Some studies also propose some building indices to identify the presence of a building. The morphological building index (MBI) \cite{huang2011multidirectional}, which takes the characteristics of buildings into consideration by integrating multiscale and multidirectional morphological operators, can be implemented to extract buildings automatically. The most widely used approaches are classification-based approaches, which make use of spectral information, structural information, and context information. The pixel shape index (PSI) \cite{zhang2006pixel}, a  shape  feature measuring the  gray  similarity  distance in each direction, is integrated with spectral  features  to extract buildings by using a support  vector  machine. However, the main problem with these algorithms is that multiple features need to be engineered for the proper classifier, which may consume too much computational resources and thus preclude large scale applications.

Based on learning data representations, deep learning is the state-of-the-art method for many big data analysis applications \cite{zhu2017deep} \cite{chen2016deep} \cite{liao2018deep} \cite{li2019deep}. Deep learning architectures such as convolutional neural networks (CNN), which is an artificial neural network based on multiple processing layers, have been extensively employed in many computer vision tasks \cite{hua2019relation} \cite{hua2019recurrently}. A major advantage of CNN is its independence from prior knowledge and hand-crafted features, which has supported its more powerful generalization capability. CNN is superior to other approaches with respect to accuracy and efficiency.  In particular, many CNN models have been proposed and applied in semantic segmentation with quite promising results, such as the fully convolutional network (FCN) \cite{long2015fully}, U-Net \cite{ronneberger2015u} SegNet \cite{badrinarayanan2017segnet}, ResNet \cite{he2016deep}, ENet \cite{paszke2016enet}, DenseNet \cite{huang2017densely}, PSPNet \cite{zhao2017pyramid}, and DeepLabv3+ \cite{chen2018encoder}. Recently, the generative adversarial network (GAN)\cite{goodfellow2014generative} has shown the potential in solving such problems.

In fact, the task of building footprint generation belongs to the branch of semantic segmentation in computer vision. In the RS community, recent research has also made an effort to improve building footprint generation through the application of the aforementioned CNN models. In order to perform building segmentation, a multi-constraint fully convolution network (MC–FCN) model is proposed in \cite{wu2018automatic}, which consists of a FCN architecture and multi-constraints. In \cite{xu2018building}, a modified and extended architecture of both ResNet and U-Net, named Res-U-Net, is proposed to improve the accuracy of building segmentation results from RS imagery. A comparatively simple and memory-efficient model, SegNet, is used for a multi-task (a shared representation for boundary and segmentation prediction) learning for building footprint generation in \cite{bischke2017multi}. A conditional GAN called cwGAN-gp \cite{shi2019building}, whose loss function is derived from the Wasserstein distance and an added gradient penalty term, is proposed to improve the building footprint generation results.

However, there are usually non-sharp boundaries and visually degraded results in CNN-based semantic segmentation tasks, which results from the inherent invariant to spatial transformations of CNN architectures. In this case, the common approach to improving the accuracy of pixel-level segmentation is to adopt a graph model such as conditional random field (CRF) as a post-processing step. Fully connected CRF \cite{kaiser2017learning} is applied to accurately localize segment boundaries and assign the most probable label to each pixel after the training based on FCN in \cite{bittner2017building}. In this case, the CRF inference is used as a post-processing step, which is not integrated with the training of the CNN. In this research, we propose an accurate and reliable building footprint generation framework, which makes three contributions:

(1)	Since each existing CNN model also has its own limitations, achieving more accurate segmentation results is still critical for automatic building footprint generation. The use of a graph model enables the combination of low-level image information such as the interactions between pixels, which is especially important for capturing fine local details. Therefore, in order to achieve more accurate segmentation results, we propose to combine CNN and a graph model in an end-to-end framework for building footprint generation, which has not been adequately addressed in the current literature. 

(2) In addition, it should be noted that, in this research, we propose a graph model called feature pairwise conditional random field (FPCRF) to be exploited in the building footprint generation framework. Specifically, we design a pairwise potential term with localized constraints in CRF. This term combines feature kernels extracted from CNN, which allows more complete feature learning than other traditional graph models. Moreover, the localized processing facilitates the efficient message passing operation.

(3)	Recently, there has been some development of deep learning methods in the computer vision community that seek to enhance the results of semantic segmentation; this development offers the RS community an opportunity to investigate the application of building footprint generation using deep learning methods. However, there is still a lack of a comprehensive investigation into the state-of-the-art CNN models in the tasks of automated building footprint generation from remote sensing imagery. With the aim of better understanding the usability and generalization ability of the state-of-the-art approaches, we compare and analyze the performances and characteristics of different CNN models for building footprint generation.

This research is organized as follows. In Section II, a brief review of related works is presented. Then, the proposed framework is introduced in Section III, followed by experiments in Section IV and results in Section V. Next, a discussion is provided in Section VI, leading to conclusions in Section VII.

\section{Related Work}
\subsection{Semantic Segmentation}Deep learning methods have been commonly used in the field of computer vision, from coarse to fine inference. Classification is the coarse inference, which makes a prediction for a whole input. Semantic segmentation is the fine-grained inference, which assigns a label to each pixel. CNN can learn an enhanced feature representation end-to-end for solving the semantic segmentation problems. FCN or encoder-decoder based architectures have been successfully implemented to produce spatially explicit label maps efficiently.

FCN is a forerunner of semantic segmentation, which transforms popular classification models to fully convolutional ones, and replaces the fully-connected layers with transposed convolutions to solve pixel labelling problems. Apart from the FCN architecture, the performance of other variants such as encoder-decoder based architectures is also remarkable. The spatial dimension has been gradually reduced with pooling layers in the encoder, while the local detail and spatial dimension are recovered in the decoder. Moreover, there are skip connections from encoder to decoder in U-Net, which makes the compensation from low-level details to high-level semantic information. In SegNet, the max-pooling indices are reused in the decoding process, which results in a substantial reduction of the number of parameters. ResNet-DUC \cite{wang2018understanding} is similar to U-Net, but uses a ResNet block instead of a normal block. In the ResNet block, the layers are reformulated as learning residual functions of the input layer, which is easier to optimize \cite{he2016deep}.  ENet consists of a large encoder and a small decoder, where the large encoder can be operated on smaller resolution data and contributes to efficient information processing. The potential of GAN is also investigated in the semantic segmentation domain. GAN comprises of two networks: a discriminator and a generator. The discriminator learns the boundary between classes, while the generator learns the distribution of individual classes. The two networks play a two-player min-max game to optimize both of their objective functions. The PSPNet is a typical example of the multi-scale processing network, which first generates a feature map from a feature extraction network (ResNet, DenseNet, etc.), and then utilizes a pyramid pooling module to combine multi-scale feature maps. DeepLab \cite{chen2017deeplab} is a state-of-the-art semantic segmentation model, which now already have 4 versions with different improvements over time: DeepLab V1, DeepLab V2, DeepLab V3 and DeepLab V3+. Both DeepLab V1 and DeepLab V2 use CRF as a postprocessing step, where the prediction could be refined both qualitatively and quantitatively. DeepLabv3 improve over previous DeepLab versions without CRF post-processing. This is due to the fact that a better way is designed to encode multi-scale context in its network architectures. DeepLabv3 is a network that does multi-scale processing, and by using altrous convolution it can achieve satisfactory results without increasing the number of parameters. The DeepLabv3+ model is a quick extension of DeepLabv3 that proposes to add an intermediate decoder module to the DeepLabv3, could recover object boundaries better. Currently, FC-DenseNet has shown superior results on terrestrial scene interpretation tasks. FC-DenseNet extends the DenseNet architecture to fully convolutional networks in pixel-level labeling tasks. In the DenseNet block, all preceding features are taken as input, and then its output features are transferred to all subsequent layers \cite{huang2017densely}. Through this feature reuse, the potential of the network can be utilized to improve the ease of training and parameter efficiency.

The development of CNN has rapidly improved the performance of semantic segmentation algorithms, which has elicited an increasing interest in the RS domain. Many research works have transferred these common CNN models and adapted them for RS imagery, which has already achieved good performance \cite{qiuFCN}. An efficient multi-scale approach is implemented for CNN in \cite{audebert2018beyond}, leveraging both a large spatial context and high resolution data to allow better semantic segmentation results. In \cite{volpi2018deep}, a multi-task learning method for semantic segmentation is proposed that learns the semantic class likelihoods and semantic boundaries across classes by CNN simultaneously. The spatial relation and channel relation modules are combined with CNN in \cite{mou2019relation}, which has achieved competitive semantic segmentation results.
 
\subsection{CNN for Building Footprint Generation}
In RS domain, semantic segmentation is often referred to in numerous applications, such as change detection \cite{mou2019learning}, land-cover classification \cite{marcos2018land}, road extraction \cite{le2018railway}, and building footprint generation \cite{vargas2019correcting} and etc. Since the building is an important object among various terrestrial targets in RS imagery, the task of building footprint generation has been heavily studied in the RS community. 

One of the CNN models commonly used for building footprint generation is FCN, which has showed superiority in accuracy as well as computational time. When applied with RS data, FCN is usually adapted. In \cite{maggiori2017convolutional}, a multiscale neuron module is designed in FCN, which is able to provide fine-grained building footprint maps. A multilayer perceptron (MLP) network is derived on top of the base FCN in \cite{maggiori2017can}, which extracts intermediate features from the base FCN to provide finer results. In \cite{bittner2018building}, three parallel FCNs are first implemented to combine different data sources, and then merged at a late stage to automatically generate a more accurate building footprint map.  A variant of FCN, which introduces an additional higher resolution skip connection, is adopted in \cite{kaiser2017learning} in order to preserve consistently improved results. The proposed method in \cite{marmanis2018classification} employs a similar strategy by adding skip connections, which can minimize information loss from downsampling.

Apart from FCN, other encoder-decoder based architectures such as SegNet is also preferred in building footprint generation, because its memory requirements are significantly lower then FCN's. In this regard, larger scale problems can be solved in parallel more efficiently at the inference stage. In \cite{yang2018building}, the building footprints across the entire continental United States are generated by SegNet with better fulfillment of the quality and computational time requirements. However, SegNet has a low edge accuracy, since it only uses a part of the layers to generate predicted output. Another encoder-decoder based architecture, U-Net, which combines both the low and high layers, is widely exploited to generate building footprint maps with their edges preserved. A Siamese U-Net \cite{ji2018fully}, where original images and their down-sampled counterparts are taken into the network separately, is proposed to improve the final results, especially for large buildings. Currently, some newly proposed networks, such as FC-DenseNet and GAN, have also demonstrated promising performances in building footprint generation. In \cite{li2018building}, a generator using FC-DenseNet and an adversarial discriminator are jointly trained for the building footprint generation from RS imagery.
\subsection{Graph Model}
Exploiting CNN for semantic segmentation tasks is still a significant challenge. The convolutional layer of CNN is a weights sharing architecture. Hence, shift invariant and spatial invariant characteristics limit spatial accuracy for segmentation tasks \cite{garcia2017review}. The convolution filters with large receptive fields and max-pooling layers in CNN also lead to coarse segmentation output, such as a non-sharp boundary and blob-like shapes \cite{zheng2015conditional}.  Moreover, CNN fails to refine local details without taking the interactions between pixels into consideration. Graph models enable modeling of interactions between pixels, which can integrate more elaborate terms to preserve the sharp boundary. Therefore, graph models can be utilized to enhance the semantic segmentation results from CNN, which has the ability to capture fine-grained details.

A graph model is a probabilistic model that encodes a distribution based on a graph-based representation. In a graph model, conditional dependencies are expressed between random variables. There are two categories of graphical representations of distributions, Bayesian networks and Markov random field (MRF), which are distinguished by their encoded set of independence and induced factorization of the distribution. In the Bayesian networks, the network structure of the model is based on a directed acyclic graph, where the joint distribution is represented as a product of conditional distributions. MRF is an undirected graph, which is described by random variables with a Markov property. In the Markov property, only the present state contributes to the conditional probability distribution of future states of the process. CRF is a notable variant of MRF, in which each random variable is conditioned upon some global observations. FullCRF \cite{zheng2015conditional} is a notable example of CRF, which is regarded as a recurrent neural network (RNN) that forms a part of a deep network for end-to-end training. However, FullCRF is based on a complex data structure and does not allow efficient GPU computation. Recently, there are some researches focused on the improvement of CRF.  The work in \cite{jampani2016learning} proposes to use bilateral convolution layers (BCL) built inside CNN architectures for efficient CRF inference, where the receptive field of filters could change. ConvCRF \cite{teichmann2018convolutional} is a recently proposed CRF algorithm that adds a conditional independence assumption to supplement FullCRF, and such an adjustment reduces the complexity of the pairwise potential. A recent example is Pixel-adaptive convolution (PAC)-CRF \cite{su2019pixel}, propose a pixel-adaptive convolution (PAC) for efficient inference of CRF to alleviate the computation, whose filter weights depends on a spatially varying kernel utilizing local pixel features.

Some researchers have tried to implement both CNN models and graph  models for building footprint generation. The results have shown that combining graph models and CNN models can lead to better results, especially along the boundaries of  buildings \cite{shi2020building}. In \cite{vakalopoulou2015building}, MRF is integrated as a post-processing stage after the training of CNN, which has ameliorated the final building footprint generation map. The CRF is exploited in \cite{paisitkriangkrai2015effective} and \cite{paisitkriangkrai2016semantic} to smooth the final pixel labeling results from CNN, which can respect the edges present in the imagery. However, the graph models are exploited only as post-processing steps in these studies. In \cite{zhuo2018optimization}, the FullCRF is plugged in at the end of the FCN for end-to-end training, which has preserved sharp boundaries, but requires longer training time and greater efforts to find optimal parameters.
\section{Methodology}
In this section, the proposed building footprint generation framework is first described. Then, we introduce the proposed FPCRF, which has a designed pairwise potential term for complete feature learning and efficient computation. The experiment design for detailed investigation of FPCRF parameters is provided in Section IV.C.

\subsection{The Proposed Building Footprint Generation Framework}
The building footprint generation in our research is actually a semantic segmentation task in the computer vision field. Recently, CNN has achieved great success in semantic segmentation tasks, as it is able to learn a strong feature representation instead of hand-crafted features. However, there are also some problems with CNN models, such as limited spatial accuracy, non-sharp boundaries, and so on. Parallel with CNN models, graph models, which enable interactions between pixels to be modeled, have also been shown to be effective methods to improve semantic segmentation results. For example, sharp boundaries and fine-grained details can be preserved by graph models. In order to harness the strengths of both models, we propose to integrate CNN and a graph model in the framework of building footprint generation. However, it should be noted that although the results could be improved by simply including graph models after learning from CNN, an end-to-end training scheme that fully integrates the graph models with CNN is preferred in our research. The end-to-end approach can provide more replicable and stable building footprint maps, especially for large scale applications. In this regard, we propose to utilize FPCRF as the graph model in the end-to-end framework, as it is superior to other graph models in terms of computation efficiency and completeness in feature learning.

\begin{figure*}[htbp]
\captionsetup{justification=centering}
  \includegraphics[width=\linewidth]{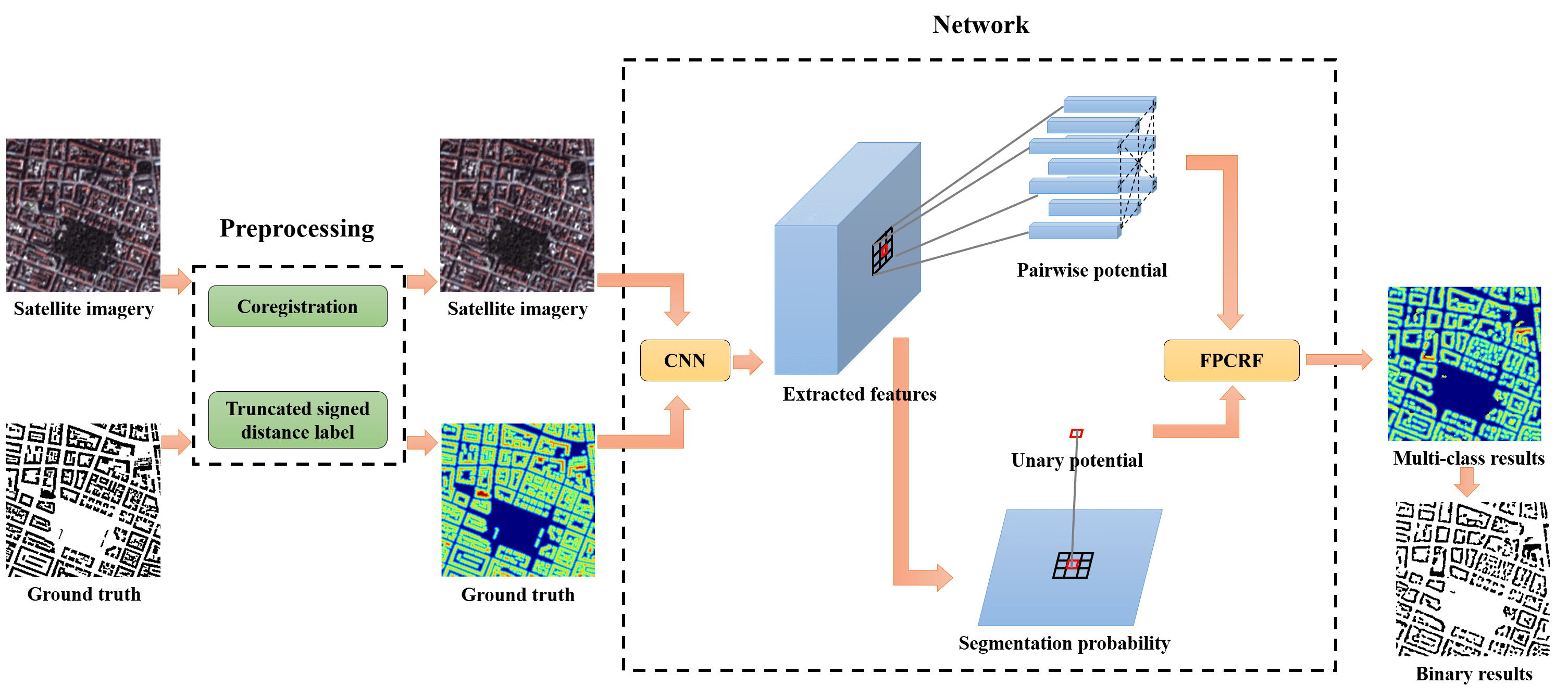}
  \caption{Flowchart of the proposed approach}
  \label{Fig. 1}
\end{figure*}

In our proposed approach, CNN and FPCRF are integrated in an end-to-end framework, where the gradients are propagated through the entire pipeline. In this case, CNN and FPCRF can co-adapt and therefore produce the optimal output. Fig. \ref{Fig. 1} shows the overall architecture of the proposed approach. It has two major components: CNN and FPCRF. The output of the CNN consists of two parts. One output is the segmentation probability obtained from the last softmax layer of CNN, which predicts labels for pixels. This segmentation probability obtained from CNN is utilized as the unary potential \cite{zheng2015conditional}. The other output is extracted features from CNN, which encodes each pixel as a fixed-length vector representation (i.e. embedding). This feature embedding is used for pairwise potential calculation, which encourages assigning similar labels to pixels with similar properties. The FPCRF component is utilized as the graph model to complement the results obtained from CNN. FPCRF takes the patch of feature embedding and unary potential as input and models their spatial correlations. The final output from FPCRF is the marginal distribution of each pixel, which represents the different class label when the patch embedding is given. 

\subsection{Data Preprocessing}
Since the ground truth of the building footprint is generated using OSM with different data sources from satellite images, the inconsistencies between datasets need to be resolved by the preprocessing steps, including coregistration and truncated signed distance labels.   

\begin{figure}
   \subfloat[]{\includegraphics[width=0.22\textwidth]{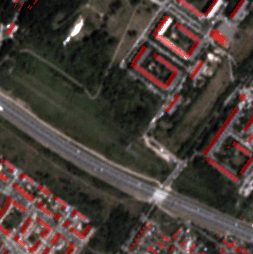}}
   \hspace*{\fill}
   \subfloat[]{\includegraphics[width=0.22\textwidth]{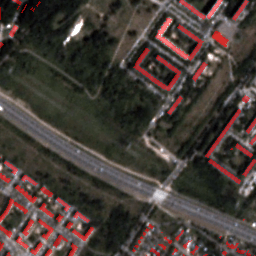}}
   \caption{(a) Before coregistration, (b) After coregistration.}
   \label{Fig. 3}
 \end{figure}

(1) Coregistration: One inconsistency is the misalignment between OSM building footprints and satellite imagery, which is caused by different projections and accuracy levels from data sources. This misalignment leads to inaccurate training samples, which need to be corrected. In this regard, we make an assumption that after translation the building footprint from OSM will be aligned with satellite imagery content within a local neighborhood \cite{yuan2014learning}. Between the building footprint and gradient magnitude of satellite imagery, a cross correlation is calculated, where the maximum of the cross-correlation corresponds to the estimated alignment location. In this regard, the offsets in both row and column direction can be derived, which are corresponding to the translation coefficients. An example of satellite imagery overlaid with the OSM building footprint is presented in Fig. \ref{Fig. 3} (a).  There are noticeable misalignments between the building footprint and the satellite imagery. The local neighborhood size is selected as 7. Fig. \ref{Fig. 3} (b) illustrates the coregistration result.

(2) Truncated signed distance label: In order to incorporate both semantic information and geometric properties of the buildings during training \cite{bischke2017multi}, the distances from pixels to the boundaries of buildings are extracted as output representations. In our experiment, the signed distance from a pixel to its closest point on the boundary is calculated with positive values indicating building interior and negative indicating building exterior. Then we truncate the distance at a given threshold to only incorporate the pixels closest to the border \cite{bischke2017multi}. Finally, the distance values are categorized into a number of class labels \cite{bischke2017multi}. The advantages of this truncated signed distance mask is that the location of the boundary and implicit geometric properties of each pixel can be captured. In addition, different buildings can be distinguished based on their between-distance and labels.

Given that $J$ is the set of pixels on the object boundary and $L_{l}$ is the set of pixels with class label $l$, the truncated distance $D(i)$ for every pixel $i$ is calculated as
\begin{equation}
\begin{aligned}
    D(i)=\delta_{p}min(min_{\forall j \in J} d_{eu}(i,j),T) \\
    \delta_{p}=\left\{ \begin{array}{ll}
1 & \textrm{if $p\in L_{building}$} \\
-1 & \textrm{if $p\in L_{non-building}$} \\
\end{array} \right.
\end{aligned}
\end{equation}
, with $d_{eu}(i,j)$ being the Euclidean distance between pixels $i$ and $j$ and $T$ is the truncated threshold. The sign function $\delta_{p}$ is used to weight the pixel distances to represent whether the pixels are inside or outside the building masks. To facilitate training, the continuous distance values are then uniformly quantized.

\begin{figure}
   \subfloat[]{\includegraphics[width=0.22\textwidth]{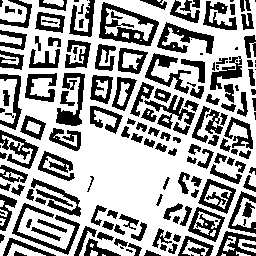}}
   \hspace*{\fill}
   \subfloat[]{\includegraphics[width=0.22\textwidth]{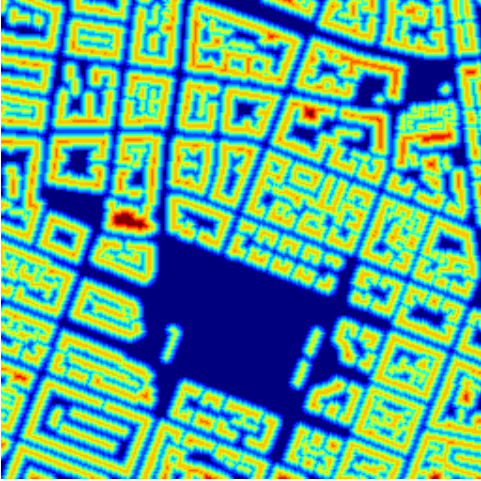}}
   \hfil
   \begin{center}
   \subfloat[]{\includegraphics[width=0.22\textwidth]{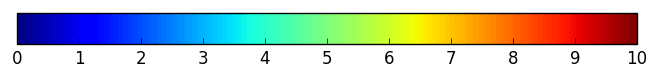}}
   \end{center}
   \caption{(a) Binary label, (b) Truncated signed distance label, (c) Colorbar for the class label. }
   \label{Fig. 4}
 \end{figure}

In this research, we use 11 classes with the labels $L=\{ 0,1,2,...,10 \}$. Class 5 represents the building boundary and when the class label is greater than 5, this pixel belongs to the building. Similarly, the non-building pixel has a class label smaller than 5. Fig. \ref{Fig. 4} illustrates the binary label and truncated signed-distance label of a building footprint, which are used in the network training. Based on the raw output (multiclass) from a trained network, we simply select a threshold to classify the class labels as a final binary building footprint result: a pixel is considered as building if $l>=5$; otherwise it is considered as non-building when $l<5$.

\subsection{FPCRF}
An image can be regarded as a graph, where every pixel is a vertex, and there are edges between each pair of pixels. FPCRF provides a probabilistic model for an image that is both local and modular.

In FPCRF, the joint probability for the random variables is implied as functions over cliques,
\begin{equation}
\begin{aligned}
    P(X=x\mid I)=\frac{1}{Z(I)}exp(-\sum_{c\in C_{G}} \phi_{c}(X_{c}\mid I))
\end{aligned}
\end{equation}
, where $X$ is a field defined over a set of variables $\{ X_{1},...,X_{N} \}$ with $N$ being the number of pixels, where the domain of each variable is a set of labels $L= \{ l_{1},l_{2},...,l_{c} \}$ with $c$ being the number of classes. The expression $G=(V,\varepsilon )$ denotes a graph where $V=\{X_{1}, X_{2},...,X_{N}\}$. The term $I=\{I_{1}, I_{2},...,I_{N}\}$ is a global observation (image). The term $\phi_{c}$ is a potential induced by the clique  $C_{G}$ (each two vertices are linked) in the graph $G$. The function $Z(I)=\sum exp(-\sum_{c\in C_{G}}\phi_{c}(X_{c}\mid I))$ is a partition function. The energy of a labeling is $E(x\mid I)= \sum_{c \in C_{G}} \phi_{c}(X_{c} \mid I)$. Gibbs distribution is a probability distribution that measures a system with a certain state as a function of that state's energy. Conditional random filed (CRF) explicitly gives a representation of the conditional independence between nodes of a graph. CRF and Gibbs distribution are proved to be equivalent with regard to the same graph from the Hammersley-Clifford theorem \cite{sutton2012introduction}, which indicates that when the Gibbs distribution is given, the conditional independence specified by the corresponding CRF will be satisfied by all of the Gibbs joint probability distributions. Therefore, the Gibbs distribution characterized by FPCRF can thus be expressed as
\begin{equation}
\begin{aligned}
    P(X=x\mid I)=\frac{1}{Z(I)}exp(-E(x\mid I))
\end{aligned}
\end{equation}

In order to take (1) the interactions between pixels, and (2) the approximation inference into consideration during learning, the Gibbs energy is expressed as
\begin{equation}
\begin{aligned}
    E(x\mid I)=\sum_{i \le N} \psi_{u}(x_{i}\mid I)+\sum_{i \ne j \le N} \psi_{p}(x_{i},x_{j}\mid I)
\end{aligned}
\end{equation}
, and $i$ and $j$ range from 1 to $N$. The term $\psi_{u}(x_{i}\mid I)$ is the unary potential, which is independent for each pixel. Unary potential is a distribution over the label assignment $x_i$ from the classifier. The term $\psi_{p}(x_{i},x_{j}\mid I)$ is a pairwise potential function that is determined based on the compatibility among pairs of pixels. This pairwise potential term can overcome the drawbacks of the noisy and inconsistent labeling from the unary potential alone.

In FPCRF, the pairwise potential $\psi_{p}(x_{i},x_{j}\mid I)$ is defined by the expression below,
\begin{equation}
\begin{aligned}
    \psi_{p}(x_{i},x_{j}\mid I)=\mu(x_{i},x_{j})\underbrace{\sum_{m=1}^M w^{(m)} k^{(m)} (f_{i},f_{j})}_{k(f_{i},f_{j})}
\end{aligned}
\end{equation}
, where $w^{(m)}$ are learnable parameters, and $M$ is the number of kernels, which is determined by the selected kernels. The terms $f_{i}$ and $f_{j}$ are feature vectors for pixels $i$ and $j$ and may depend on the input image $I$. The function $\mu(x_{i},x_{j})$ is the compatibility transformation and captures the compatibility between labels $x_{i}$ and $x_{j}$.

However, FullCRF and ConvCRF only use shallow features --- the color and position of the pixel for kernels in pairwise potential term, which have not fully harnessed the complete features extracted from CNN. In this regard, we propose FPCRF as a graph model to be exploited in the building footprint generation framework. 

Inspired by the fact that ConvCRF is based on localized processing, we design a pairwise potential term with localized constraints in FPCRF that allows complete feature learning. The kernel utilized for pairwise potential in FPCRF is a Gaussian kernel, which is defined by the feature vectors $f_{1}$, ... , $f_{B}$, where $B$ is the number of feature vector types. The kernel $k^{(m)}$ is defined as:
\begin{equation}
\begin{aligned}
    k^{(m)}(f_{i},f_{j})=exp(-\sum_{b=1}^{B}\frac{|f_{b,i}-f_{b,j}|^2}{2\theta_{b}^2})
\end{aligned}
\end{equation}
, where $\theta_{b}$ is a learnable parameter.

The labeling of the random field is derived by the maximum a posteriori (MAP) method,
\begin{equation}
\begin{aligned}
x^{*}=argmax_{x \in L^{N}}P(X=x \mid I)
\end{aligned}
\end{equation}

The most probable label $x$ can be yielded by the minimization of the Gibbs energy in FPCRF. However, the exact minimization is intractable. In this regard, the mean field inference is utilized for the approximation of FPCRF distribution. A distribution $Q(X)$ that tries to minimize the KL-divergence $D (Q||P)$ from exact distribution $P(X)$ is computed by the mean field approximation,
\begin{equation}
\begin{aligned}
  D(Q||P)=\sum_{x}Q(x)log(\frac{P(x)}{Q(x)}),
\end{aligned}
\end{equation}
, where the approximated distribution $Q(X)$ can be represented as a product of independent marginal distributions,
\begin{equation}
\begin{aligned}
    Q(X)=\prod_{i}Q_{i}(X_{i})
\end{aligned}
\end{equation}

 The combined message passing result $Q$ of all kernels is expressed as:
\begin{equation}
\begin{aligned}
    Q_{i}(x_{i}=l)=\frac{1}{Z_{i}}exp\{-\psi_{u}(x_{i}\mid I)\\-\sum_{l'\in L}\mu(l,l')\sum_{m=1}^M w^{(m)} \sum_{d_{ma}(i,j)<r}k^{(m)} (f_{i},f_{j}) Q_{j}(l')\}
\end{aligned}
\end{equation}
 
The steps of the mean field algorithms are presented in Table \ref{Tab. 0}.

\begin{table*}[htbp]
  \captionsetup{justification=centering}
  \caption{The steps of the mean field algorithms in FPCRF}
\begin{center}
\begin{tabular}{ll}
  \hline\hline
   Mean field approximation in FPCRF & \\
     1. Initialize Q     &   ${Q}_{i}(x_{i})=\frac{1}{Z_{i}}exp\{-\psi_{u}(x_{i}\mid I)\}$ \\
     2. while not converged & \\
     3. $\tilde{Q}_{i}^{(m)}(l)\leftarrow \sum_{i \neq j}k^{(m)} (f_{i},f_{j}) Q_{j}(l)$ for all m & Message passing from all $X_{j}$ to all $X_{i}$ \\
     4. $\check{Q}_{i}(x_{i})\leftarrow \sum_{m=1}w^{(m)}\tilde{Q}_{i}^{(m)}(l)$ & Weighting filtering outputs \\
     5. $\hat{Q}_{i}(x_{i})\leftarrow \sum_{l'\in L}\mu(l,l')\check{Q}_{i}(l)$ & Compatibility transformation \\
     6. $Q_{i}(x_{i})\leftarrow -\psi_{u}(x_{i}\mid I)-\hat{Q}_{i}(x_{i})$ & Adding unary potentials \\
     7. $Q_{i}(x_{i})\leftarrow \frac{1}{Z_{i}}exp(-Q_{i}(x_{i}))$ & Normalization \\
     8. end while & \\
     \hline
\end{tabular}
\end{center}
\label{Tab. 0}
\end{table*} 

The steps of the mean field inference algorithm of FPCRF are reformulated as a network layer, where the error differentials in each layer with respect to its inputs are sent to previous layers by back propagation during training  \cite{zheng2015conditional}. FPCRF exploits a $1 \times 1$ filter to assign the different penalties for all different pairs of labels. 

To implement the efficient computation of the convolution, the input is firstly tiled into the specific shapes, which are related to the filter size $r$. An efficient message passing operation in FPCRF can be implemented analogously to 2D-convolution \cite{teichmann2018convolutional}. Then, the message passing step is reformulated to be a convolution with a truncated Gaussian kernel. 

\section{Experiments}
\subsection{Study Area and Dataset}
\begin{figure*}[htbp]
\captionsetup{justification=centering}
  \includegraphics[width=\linewidth]{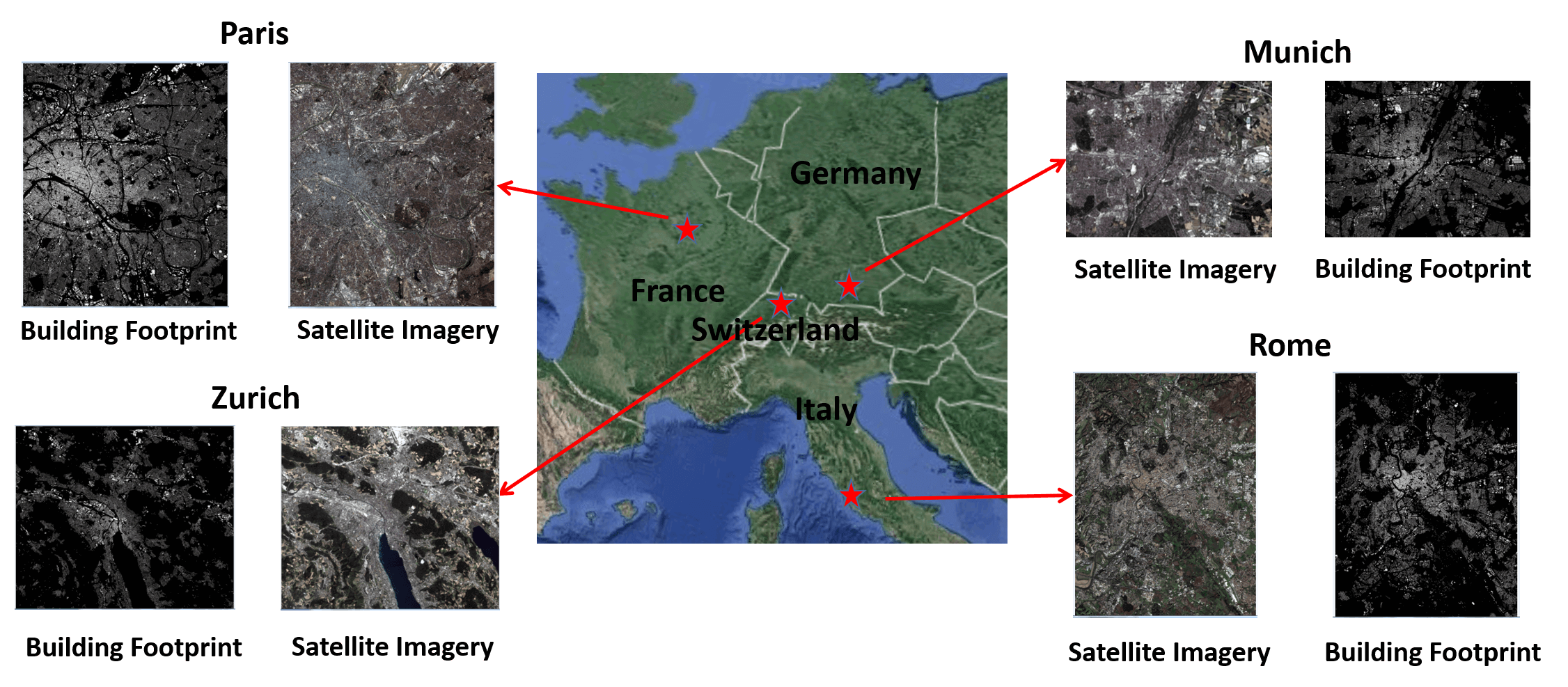}
  \caption{True color Planetscope satellite images and building footprint of Munich, Rome, Paris, and Zurich}
  \label{Fig. 2}
\end{figure*}

In this research, the study sites cover four cities (see Fig. \ref{Fig. 2}): (1) Munich, Germany; (2) Rome, Italy; (3) Paris, France; (4) Zurich, Switzerland. We use Planetscope satellite imagery \cite{planet} with three bands --- Red, Green, Blue (RGB) --- and 3 m spatial resolution to validate our proposed method. The imagery is processed using a $256 \times 256$ sliding window. The corresponding building footprint (stored as polygon shape files) is downloaded from OSM, where the detailed building footprints around these four cities are publicly released. Some patches are mismatched, which result from the time difference between OSM building footprints and satellite imagery. For example, a building might appear in the OSM building footprint, while it is missing in the corresponding satellite imagery, or vice versa. To limit such patches, we have manually selected 3000 pairs of proper patches. The selected pairs are then separated into two parts, where 80\% of the sample patches are used for training the network and 20\% are used for  model validation. 

\subsection{Experiment Setup}
In this research, all networks were investigated within a Pytorch framework on an NVIDIA Titan X GPU with 12 GB of memory \cite{JUWELS}. For all networks, a stochastic gradient descent (SGD) optimizer with a learning rate of 0.0001 was utilized and negative log likelihood loss (NLLLoss) was taken as loss function. The batch size of all networks  was 4.

In our proposed end-to-end approach, CNN and FPCRF are two vital parts in the framework, where the CNN component acts as a feature extractor, and the FPCRF models their pixel correlations by using pairwise potential. Hence, we first investigate which CNN model has better feature extraction capability. Then, the feature kernels that are taken in pairwise potential calculation of FPCRF are also carefully studied to find the optimal feature embedding. Moreover, the sensitivity of the filter size $r$, being the only hyperparameter of FPCRF, is analyzed. Additionally, to prove the superiority of our proposed framework, we train the following networks for comparison:

1)	FCN-8s is based on VGG16 as the encoder and an up-sampling layer and convolutional layer as the decoder.

2)	ResNet-DUC, which has [3, 4, 6, 3, 3, 6, 4, 3] convolutional layers in each ResNet block.

3)	SegNet, which attaches a reversed VGG16 as a decoder to the encoder.

4)	U-Net, which has a depth of five with a feature channel in each depth [64, 128, 256, 512, 1024].

5)	ENet, which consists of five stages, where the first three stages act as the encoder, while the last two stages belong to the decoder.

6)	cwGAN-gp which also has five depth U-Net in the generator.

7)	FC-DenseNet, with each dense block having [5, 5, 5, 5, 5, 5, 5, 5, 5, 5, 5, 5] convolutional layers.

8) PSPNet starts off with a standard feature  extraction network (ResNet101).

9) DeepLabv3+ utilizes the Xception model \cite{chollet2017xception} as the feature extractor.

\section{Results}
The three metrics in the following experiments selected to evaluate the results are overall accuracy, F1 score, and intersection over union (IoU), which are used widely to evaluate building footprint generation results.
\begin{equation}
\begin{aligned}
\textrm{Overall accuracy}=\frac{TP+TN}{TP+FP+FN+TN}
\end{aligned}
\end{equation}
\begin{equation}
\begin{aligned}
\textrm{precision}=\frac{TP}{TP+FP}
\end{aligned}
\end{equation}
\begin{equation}
\begin{aligned}
\textrm{recall}=\frac{TP}{TP+FN}
\end{aligned}
\end{equation}
\begin{equation}
\begin{aligned}
\textrm{F1 score}=\frac{2*\textrm{precision}*\textrm{recall}}{\textrm{precision}+\textrm{recall}}
\end{aligned}
\end{equation}
\begin{equation}
\begin{aligned}
\textrm{IoU}=\frac{TP}{TP+FP+FN}
\end{aligned}
\end{equation}
, where $TP$ is the number of building pixels correctly detected, and $FN$ denotes the missed building pixels. $FP$ and $TN$ are the numbers of non-building pixels in the ground reference, but detected as buildings and non-buildings in the result, respectively. The F1 score indicates a balance between precision and recall.

\subsection{Feature Extractor Combined with FPCRF}
\begin{figure*}
   \subfloat[]{\includegraphics[width=0.22\textwidth]{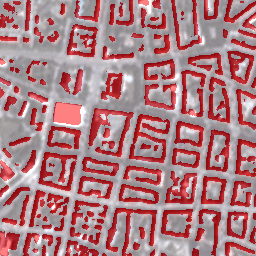}}
   \hspace*{\fill}
   \subfloat[]{\includegraphics[width=0.22\textwidth]{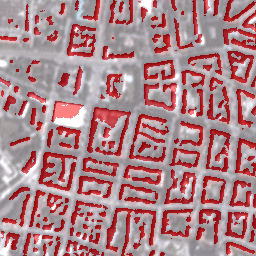}}
   \hspace*{\fill}
   \subfloat[]{\includegraphics[width=0.22\textwidth]{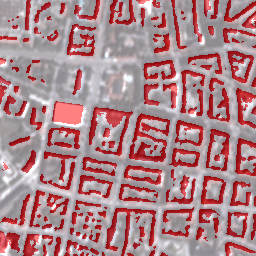}}
   \hspace*{\fill}
   \subfloat[]{\includegraphics[width=0.22\textwidth]{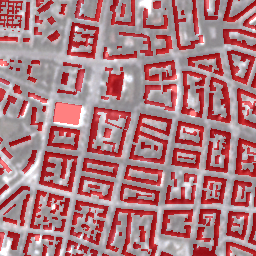}}
   \caption{The predicted results (in red) obtained from different feature extractors (a) FC-DenseNet, (b) FCN-8s, (c) U-Net combined with FC-DenseNet, and (d) ground truth. }
   \label{Fig. 8}
 \end{figure*}

Fig. \ref{Fig. 8} and Table \ref{Tab. 3} list the results of the different CNN models combined with FPCRF. The results of FC-DenseNet combined with FPCRF are more accurate than other two CNN models combined with FPCRF. This is due to the superiority of FC-DenseNet, which extends the DenseNet architecture to FCN for semantic segmentation. In the DenseNet block, through feature reuse, there are shorter connections within the layers close to the input or output, which strengthen the learning of the discriminated features. Moreover, features are combined by iterative concatenation, which contributes to the improved flow of information. In addition, a standard skip connection between the encoder and decoder is used to pass higher resolution information, which can help the encoder recover spatially detailed information from the decoder.

\begin{table}[htbp]
\captionsetup{justification=centering}
\caption{Accuracy of different feature extractors combined with FPCRF }
\begin{center}
\setlength\tabcolsep{2.5pt}
\begin{tabular}{llll}
  \hline\hline
   Methods & Overall accuracy & F1 score & IOU\\
   \hline\hline
    \begin{bfseries}FC-DenseNet + FPCRF\end{bfseries} & \begin{bfseries}0.9297\end{bfseries}	& \begin{bfseries}0.6698\end{bfseries} & \begin{bfseries}0.5046\end{bfseries} \\
    FCN-8s + FPCRF & 0.9248	& 0.6340 & 0.4642 \\
    U-Net+ FPCRF & 0.8927	& 0.6278 & 0.4575 \\
     \hline
\end{tabular}
\end{center}
\label{Tab. 3}
\end{table}

\subsection{Kernel Selection in FPCRF}
\begin{figure*}
   \subfloat[]{\includegraphics[width=0.22\textwidth]{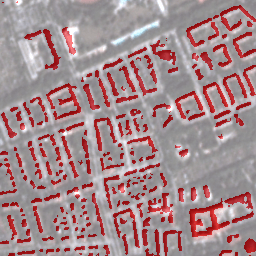}}
   \hspace*{\fill}
   \subfloat[]{\includegraphics[width=0.22\textwidth]{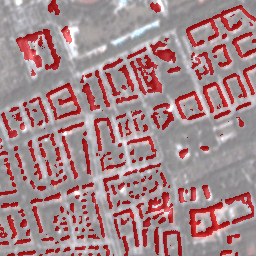}}
   \hspace*{\fill}
   \subfloat[]{\includegraphics[width=0.22\textwidth]{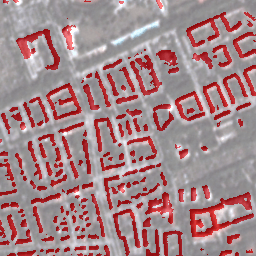}}
   \hspace*{\fill}
   \subfloat[]{\includegraphics[width=0.22\textwidth]{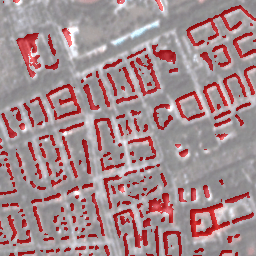}}
   \hfil
   \subfloat[]{\includegraphics[width=0.22\textwidth]{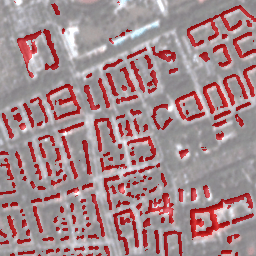}}
   \hspace*{\fill}
   \subfloat[]{\includegraphics[width=0.22\textwidth]{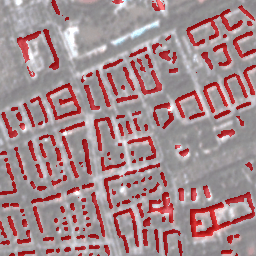}}
   \hspace*{\fill}
   \subfloat[]{\includegraphics[width=0.22\textwidth]{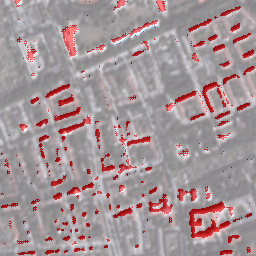}}
   \hspace*{\fill}
   \subfloat[]{\includegraphics[width=0.22\textwidth]{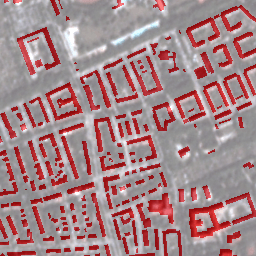}}
   \caption{The predicted results (in red) obtained from FC-DenseNet combined with FPCRF from different kernels: (a) a+s, (b) a+s+fd, (c) s+fd, (d) a+fd, (e) fs, (f) fd, (g) fc, (h) ground truth. (a: appearance kernel, s: smooth kernel, fd: feature difference kernel, fs: feature spatial kernel, fc: feature cosine kernel) }
   \label{Fig. 7}
 \end{figure*}

FullCRF and ConvCRF only utilize the  pairwise potentials from shallow features, which include only appearance and smooth Gaussian kernels. In the implementation of ConvCRF, the unary potential is obtained from CNN, and only the smooth kernel and appearance kernel are utilized for the calculation of the pairwise potential term. FPCRF is able to reduce the complexity of the pairwise potential greatly, which makes the exact message passing and complete feature learning possible. In this regard, we can use the features extracted from CNN models to calculate pairwise potentials, which may facilitate training. The results for the FC-DenseNet combined with FPCRF from the different kernels  $k^{(m)}(f_i, f_j)$ are presented in Table \ref{Tab. 2} and Fig. \ref{Fig. 7}. The appearance kernel (a) and the smooth kernel (s) are the same as FullCRF and ConvCRF. The feature difference kernel (fd) represents the CNN extracted feature difference calculated with a Gaussian function, and the feature spatial kernel (fs) is the feature difference combined with position difference calculated with a Gaussian function. In the feature cosine kernel (fc), the cosine distance between feature vectors is implemented as pairwise potential \cite{li2018cancer}. The detailed formulas of the different kernels are listed in the below:

(1) appearance kernel (a):
\begin{equation}
\begin{aligned}
    k^{(a)}(f_{i},f_{j})=exp(-\frac{|f_{p,i}-f_{p,j}|^2}{2\theta_{\alpha}^2}-\frac{|f_{I,i}-f_{I,j}|^2}{2\theta_{\beta}^2})
\end{aligned}
\end{equation}
, where $f_p$ is the feature of position, $f_I$ is the feature of color, $\theta_{\alpha}$ and $\theta_{\beta}$ are learnable parameters.

(2) smooth kernel (s):
\begin{equation}
\begin{aligned}
    k^{(s)}(f_{i},f_{j})=exp(-\frac{|f_{p,i}-f_{p,j}|^2}{2\theta_{\gamma}^2})
\end{aligned}
\end{equation}
, where $\theta_{\gamma}$ is a learnable parameter.

(3) feature difference kernel (fd):
\begin{equation}
\begin{aligned}
    k^{(f_d)}(f_{i},f_{j})=exp(-\frac{|f_{f,i}-f_{f,j}|^2}{2\theta_{\delta}^2})
\end{aligned}
\end{equation}
, where $f_f$ is the feature extracted from CNN and $\theta_{\delta}$ is a learnable parameter.

(4) feature spatial kernel (fs):
\begin{equation}
\begin{aligned}
    k^{(f_s)}(f_{i},f_{j})=exp(-\frac{|f_{f,i}-f_{f,j}|^2}{2\theta_{\zeta}^2}-\frac{|f_{p,i}-f_{p,j}| ^2}{2\theta_{\eta}^2})
\end{aligned}
\end{equation}
, where $\theta_{\zeta}$ and $\theta_{\eta}$ are learnable parameters.

(5) feature cosine kernel (fc):
\begin{equation}
\begin{aligned}
    k^{(f_c)}(f_{i},f_{j})=(1-\frac{|f_{f,i}\cdot f_{f,j}|^2}{\|f_{f,i}\|\|f_{f,j}\|})
\end{aligned}
\end{equation}

“FC-DenseNet + FPCRF (a+s)” is corresponding to the “ConvCRF ”, which means that unary potential is the segmentation probability obtained from FC-DenseNet, but for the calculation of the pairwise potential term only the smooth kernel and appearance kernel are utilized. It should be noted that in our proposed method “FC-DenseNet + FPCRF (fd)”, FC-DenseNet not only provide the segmentation probability as unary potential, but also extracts features for the calculation of the pairwise potential term. FC-DenseNet combined with FPCRF using the feature difference kernel (fd) outperforms other kernels in terms of their high F1 score and IoU. There are several reasons for this. The smooth kernel (s), which removes small isolated regions, is not useful in our case. Since the spatial resolution of satellite imagery is coarse, we can preserve isolated small buildings by removing smooth kernel. The feature spatial kernel (fs) controls the degree of nearness that neighboring pixels having similar features may belong to the same class. However, since we have already used filter size to add a locality by filter size, we want the pixels within the filter to contribute equally to the centered pixel. In addition, the appearance kernel (a) has not shown any improvements to the results. This may result from the fact that the RGB information in the appearance kernel (a) is not sufficient to distinguish the buildings from other non-building areas (sometimes roads and buildings have similar RGB information). The feature cosine kernel (fc) shows very low accuracy, which can be explained by the fact that a Gaussian function in feature difference (fd) can remove the noise, but cosine distance can be largely affected by the noise. In this case, when the cosine distance between feature vectors is implemented as a pairwise potential, the final results will suffer from great instability.

\begin{table}[h]
\captionsetup{justification=centering}
\caption{Accuracy of FC-DenseNet combined with FPCRF from different kernels. (a: appearance kernel, s: smooth kernel, fd: feature difference kernel, fs: feature spatial kernel, fc: feature cosine kernel) }
\begin{center}
\setlength\tabcolsep{2.5pt}
\begin{tabular}{llll}
  \hline\hline
   Methods & Overall accuracy & F1 score & IOU\\
   \hline\hline
    FC-DenseNet + FPCRF (a+s) & 0.9075	& 0.6653 & 0.4986 \\
    FC-DenseNet + FPCRF (a+s+fd) & 0.9166	& 0.6682 & 0.5018 \\
    FC-DenseNet + FPCRF (s+fd) & 0.9013	& 0.6660 & 0.4991 \\
    FC-DenseNet + FPCRF (a+fd) & 0.9212	& 0.6685 & 0.5013 \\
    FC-DenseNet + FPCRF (fs) & 0.9275	& 0.6673 & 0.5006 \\
    \begin{bfseries}FC-DenseNet + FPCRF (fd)\end{bfseries} & \begin{bfseries}0.9297\end{bfseries}	& \begin{bfseries}0.6698\end{bfseries} & \begin{bfseries}0.5046\end{bfseries} \\
    FC-DenseNet + FPCRF (fc) & 0.7888	& 0.4521 & 0.2921 \\
     \hline
\end{tabular}
\end{center}
\label{Tab. 2}
\end{table}

\subsection{Hyperparameter Analysis in FPCRF}
\begin{figure*}
   \subfloat[]{\includegraphics[width=0.22\textwidth]{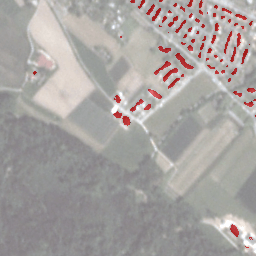}}
   \hspace*{\fill}
   \subfloat[]{\includegraphics[width=0.22\textwidth]{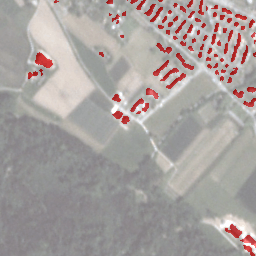}}
   \hspace*{\fill}
   \subfloat[]{\includegraphics[width=0.22\textwidth]{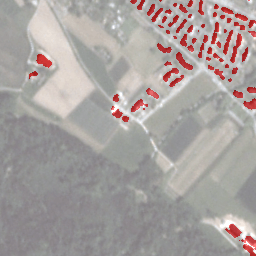}}
   \hspace*{\fill}
   \subfloat[]{\includegraphics[width=0.22\textwidth]{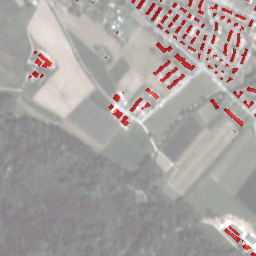}}
   \caption{The predicted results (in red) obtained from FC-DenseNet combined with FPCRF within different filter size: (a) FC-DenseNet+FPCRF ($r$=5), (b) FC-DenseNet+FPCRF ($r$=7), (c) FC-DenseNet+FPCRF ($r$=9), and (d) ground truth. }
   \label{Fig. 9}
 \end{figure*}

The hyperparameter filter size $r$ in FPCRF implies that the pairwise potential is zero when the  Manhattan distance between the pairs of pixels exceeds $r$. In order to better understand the influence of the various filter sizes $r$ for building footprint generation, the visual results of FC-DenseNet combined with FPCRF within different filter size r, as well as their accuracy indexes, are shown and compared in Fig. \ref{Fig. 9} and Table \ref{Tab. 4}. From the visual results, we can observe that when the filter size is not optimal, there are more non-building areas wrongly detected as building areas, and some small buildings are not detected. This can be explained by the fact that filter size is related to the quantity of the most useful neighboring pixels, which contributes to the improvement of the segmentation results.

\begin{table}[htbp]
  \captionsetup{justification=centering}
  \caption{Accuracy of FC-DenseNet combined with FPCRF within different filter size. $r$ is filter size.}
  \begin{center}
  \setlength\tabcolsep{2.5pt}
\begin{tabular}{llll}
  \hline\hline
   Methods & Overall accuracy & F1 score & IOU\\
   \hline\hline
    FC-DenseNet + FPCRF (r=5) & 0.9121	& 0.6665 & 0.4985 \\
    \begin{bfseries}FC-DenseNet + FPCRF (r=7)\end{bfseries} & \begin{bfseries}0.9297\end{bfseries}	& \begin{bfseries}0.6698\end{bfseries} & \begin{bfseries}0.5046\end{bfseries} \\
    FC-DenseNet + FPCRF (r=9) & 0.9142	& 0.6670 & 0.4993 \\
     \hline
\end{tabular}
\end{center}
\label{Tab. 4}
\end{table}

\section{Discussion}
\subsection{Additional Datasets}
Another three datasets, ISPRS benchmark data, Dstl Kaggle dataset, and Inria Aerial Image Labeling data are used to test the performance and characteristics of the different networks for building footprint generation.

\begin{figure}[htbp]
\captionsetup{justification=centering}
\begin{center}
  \includegraphics[width=0.5\textwidth]{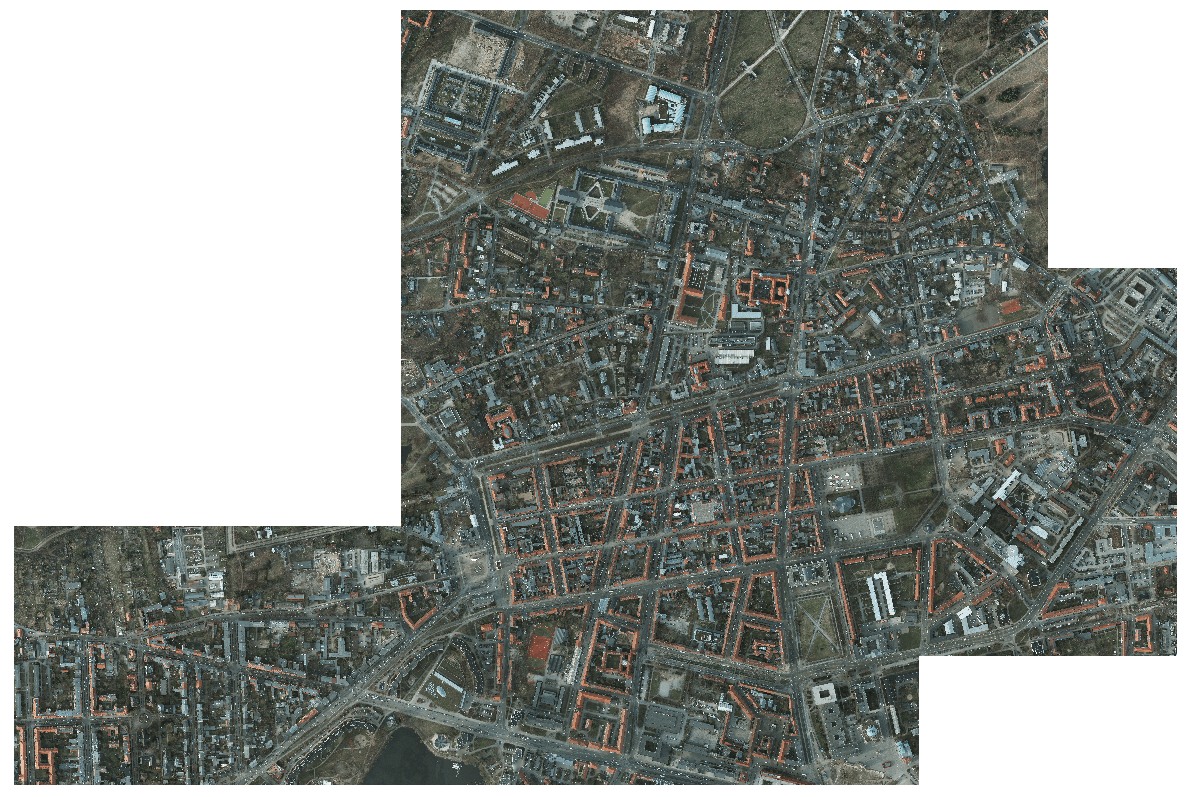}
\end{center}
  \caption{The aerial imagery in ISPRS dataset (spatial resolution: 5cm).}
  \label{Fig. 10}
\end{figure}

The first dataset is ISPRS benchmark data \cite{ISPRSdata}, shown in Fig. \ref{Fig. 10}. The dataset covers the city of Potsdam, which contains 38 aerial images with pixel size $6000 \times 6000$ and four channels: Red, Green, Blue (RGB) and Near-infrared bands with 5 cm spatial resolution. The corresponding ground truth is also available from the ISPRS benchmark data, which includes six categories. In this research, we take the building class as building and other five classes as non-building; traditional natural color aerial imagery are utilized. The images 7-07, 7-08, 7-09, 7-10, 7-11, 7-12, and 7-13 are used as the validation set, and the remaining images are exploited for training.

\begin{figure}[htbp]
\captionsetup{justification=centering}
\begin{center}
  \includegraphics[width=0.3\textwidth]{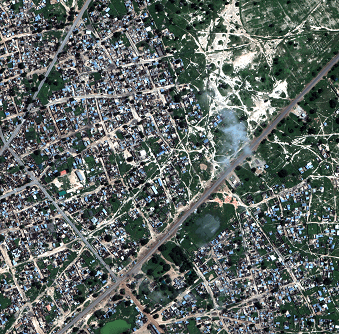}
\end{center}
  \caption{The WorldView 3 imagery in Dstl dataset (spatial resolution: 1.24m).}
  \label{Fig. 11}
\end{figure}

Dstl Kaggle dataset \cite{dstldata} is the second dataset, which provides 57 satellite images with a region of $1 km \times 1 km$ in both 3-band RGB and 16-band multi-spectral formats. Here, we use 3-band images with the spatial resolution 1.24 m. In this dataset, 10 different classes have been labeled within some images. In this research, the pixels of building are from building class, and those of non-building are the remaining pixels. Ten satellite images with pixel size $3348 \times 3348$, which has corresponding building class in the ground truth, are exploited for this experiment, which includes eight images with ID (6100-2-3, 6100-1-2, 6100-3-1, 6110-4-0, 6120-2-0, 6120-2-2, 6140-1-2, 6140-3-1) for training, and two images with ID (6100-1-3, 6100-2-2) for validation. Fig. \ref{Fig. 11} illustrates one satellite imagery sample.

\begin{figure}[htbp]
\captionsetup{justification=centering}
\begin{center}
  \includegraphics[width=0.3\textwidth]{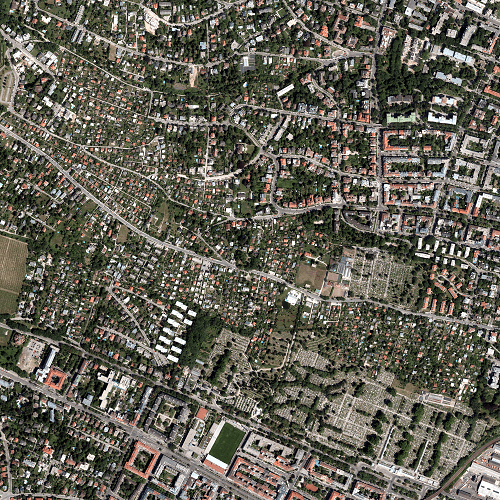}
\end{center}
  \caption{The aerial imagery in Inria dataset (spatial resolution: 30cm).}
  \label{Fig. 12}
\end{figure}

The third dataset is Inria Aerial Image Labeling data \cite{maggiori2017dataset}. This dataset contains 360 aerial images of size $5000 \times 5000$ (at a 30cm spatial resolution), which have three bands: Red, Green, and Blue. In this research, 36 tiles of aerial imagery and their corresponding ground truth (building and non-building) are selected for each of the following five regions: Austin, Chicago, Kitsap County, Western Tyrol and Vienna, where dissimilar urban settlements are covered. The sample data are showed in Fig. \ref{Fig. 12}. To split the training set and test set, we used the first eight images of every city for validation.

In order to get more training data, satellite imagery and their corresponding ground truth from Dstl Kaggle dataset are cut into small patches of size $256 \times 256$ pixels with overlap of 64. However, since the numbers of samples from ISPRS benchmark data and Inria Aerial Image Labeling data are enough for network training, aerial imagery and their corresponding ground truth from both datatsets are cut into non-overlapping patches with size $256 \times 256$ pixels. The numbers of training and validation patches for the additional three datasets are listed in Table \ref{Tab. a}.

\begin{table*}[htbp]
  \captionsetup{justification=centering}
  \caption{The numbers of training and validation patches of three additional dataset}
  \begin{center}
  \setlength\tabcolsep{2.5pt}
\begin{tabular}{llll}
  \hline\hline
   Dataset & Number of training patches & Number of validation patches\\
   \hline\hline
    ISPRS benchmark data (spatial resolution: 5cm) & 16000	& 3573 \\
    Dstl Kaggle dataset (spatial resolution: 1.24m) & 2312 & 578\\
    Inria Aerial Image Labeling data (spatial resolution: 30cm)& 50540	& 14440 \\
     \hline
\end{tabular}
\end{center}
\label{Tab. a}
\end{table*}

\subsection{Comparison with Other Models}

\begin{figure*}
   \subfloat[]{\includegraphics[width=0.22\textwidth]{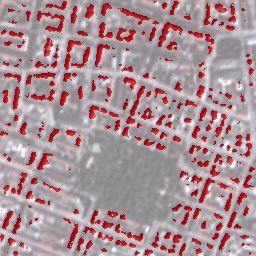}}
   \hspace*{\fill}
   \subfloat[]{\includegraphics[width=0.22\textwidth]{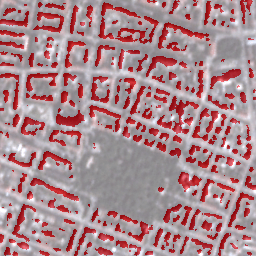}}
   \hspace*{\fill}
   \subfloat[]{\includegraphics[width=0.22\textwidth]{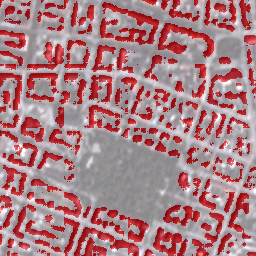}}
      \hspace*{\fill}
   \subfloat[]{\includegraphics[width=0.22\textwidth]{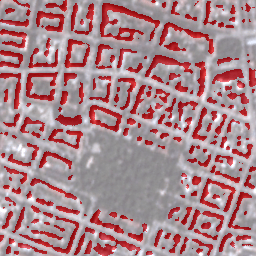}}

   \hfil
   \subfloat[]{\includegraphics[width=0.22\textwidth]{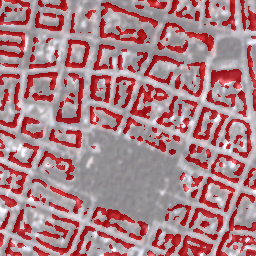}}
   \hspace*{\fill}
   \subfloat[]{\includegraphics[width=0.22\textwidth]{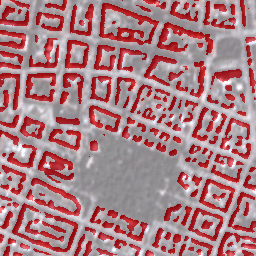}}
   \hspace*{\fill}
   \subfloat[]{\includegraphics[width=0.22\textwidth]{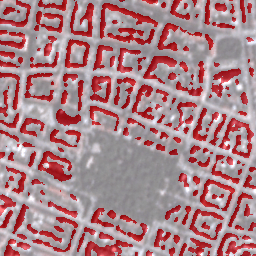}}
      \hspace*{\fill}
    \subfloat[]{\includegraphics[width=0.22\textwidth]{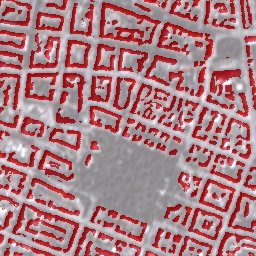}}
   \hfil
   \subfloat[]{\includegraphics[width=0.22\textwidth]{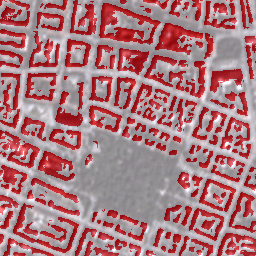}}
   \hspace*{\fill}
   \subfloat[]{\includegraphics[width=0.22\textwidth]{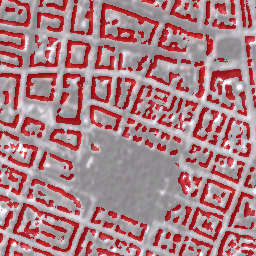}}
   \hspace*{\fill}
   \subfloat[]{\includegraphics[width=0.22\textwidth]{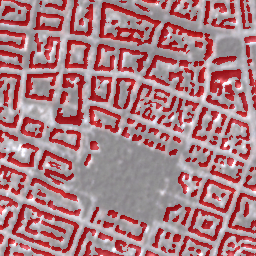}}
  \hspace*{\fill}
   \subfloat[]{\includegraphics[width=0.22\textwidth]{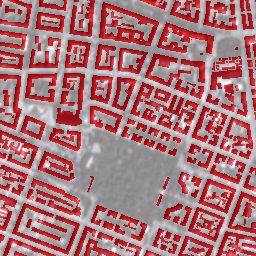}}
   \caption{The predicted results (in red) obtained from (a) ResNet-Duc, (b) SegNet, (c) ENet,  (d) U-Net, (e) FCN-8s, (f) cwGAN-gp, (g) PSPNet, (h) DeepLabv3+, (i) FC-DenseNet, (j) FC-DenseNet+FullCRF, (k) FC-DenseNet+FPCRF, and (l) ground truth from Planetscope dataset (spatial resolution: 3m).}
   \label{Fig. 13}
 \end{figure*}

\begin{figure*}
    \subfloat[]{\includegraphics[width=0.22\textwidth]{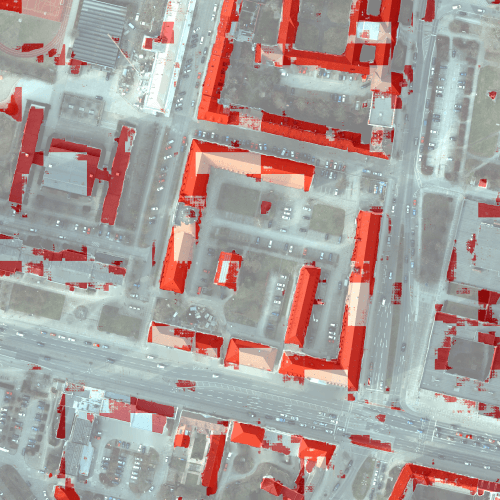}}
    \hspace*{\fill}
    \subfloat[]{\includegraphics[width=0.22\textwidth]{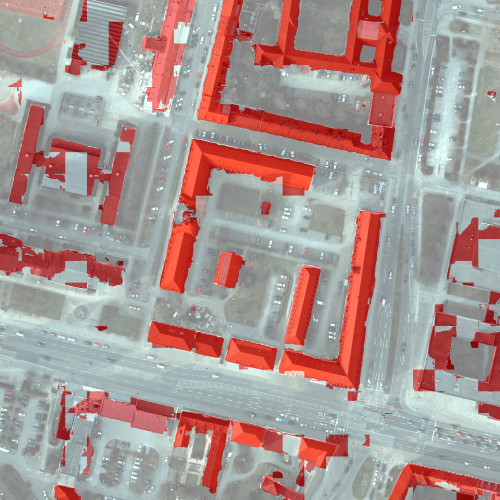}}
    \hspace*{\fill}
    \subfloat[]{\includegraphics[width=0.22\textwidth]{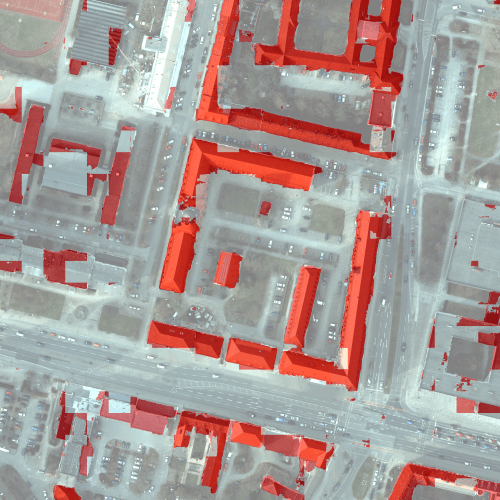}}
    \hspace*{\fill}
    \subfloat[]{\includegraphics[width=0.22\textwidth]{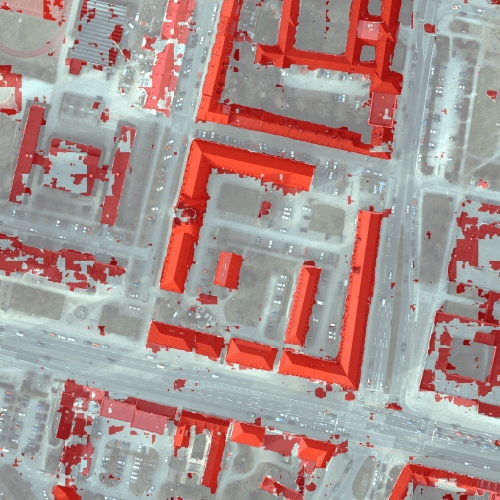}}
    \hfil
    \subfloat[]{\includegraphics[width=0.22\textwidth]{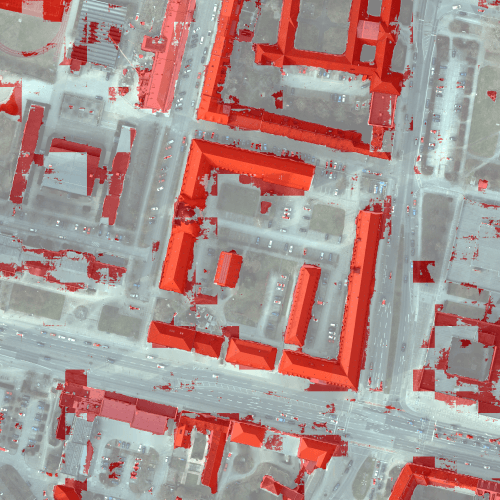}}
    \hspace*{\fill}
    \subfloat[]{\includegraphics[width=0.22\textwidth]{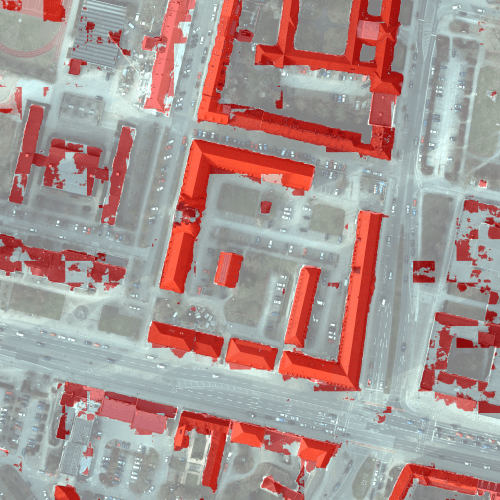}}
    \hspace*{\fill}
    \subfloat[]{\includegraphics[width=0.22\textwidth]{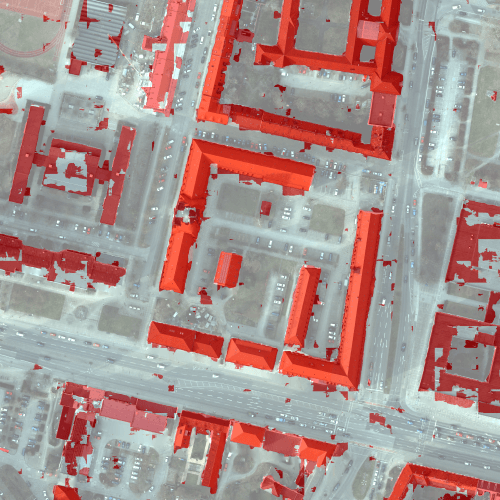}}
    \hspace*{\fill}
    \subfloat[]{\includegraphics[width=0.22\textwidth]{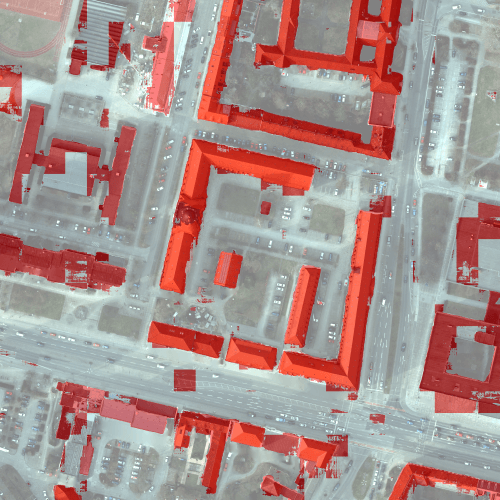}}
    \hfil
    \subfloat[]{\includegraphics[width=0.22\textwidth]{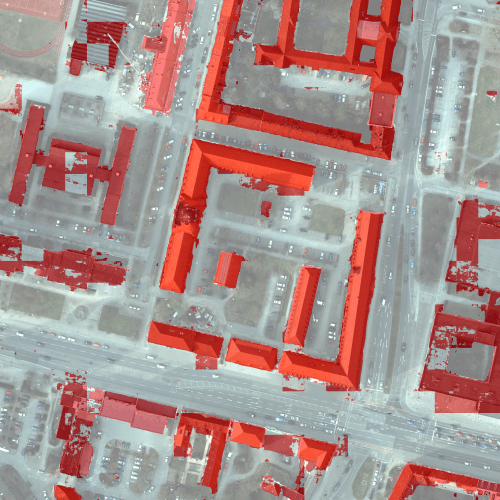}}
    \hspace*{\fill}
    \subfloat[]{\includegraphics[width=0.22\textwidth]{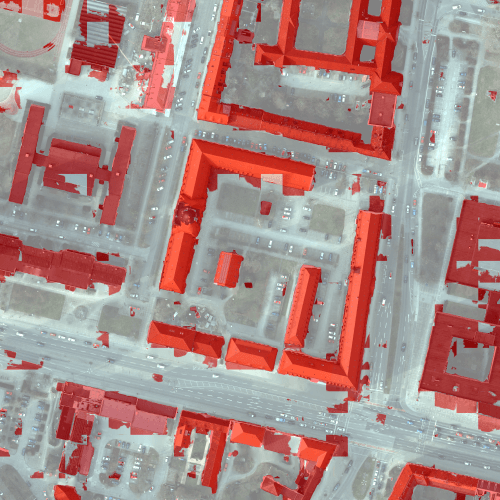}}
    \hspace*{\fill}
    \subfloat[]{\includegraphics[width=0.22\textwidth]{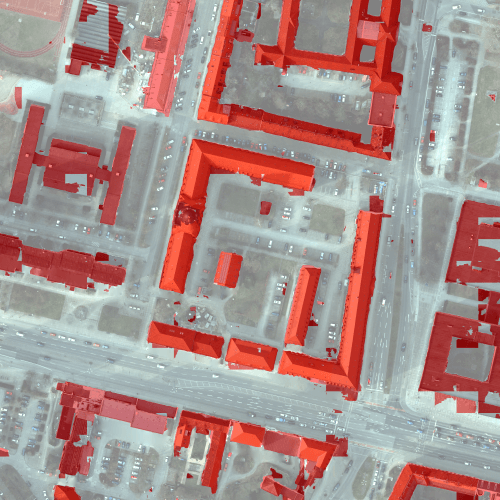}}
    \hspace*{\fill}
    \subfloat[]{\includegraphics[width=0.22\textwidth]{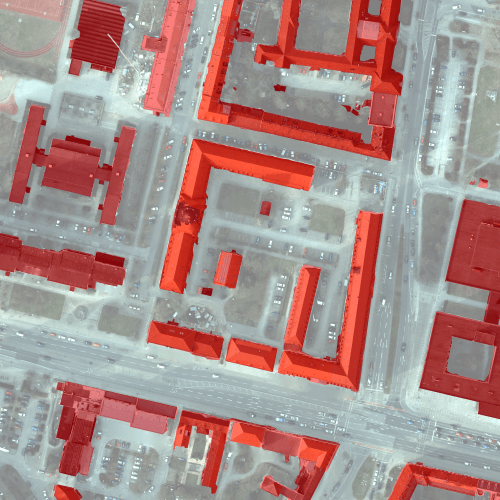}}
    \caption{The predicted results (in red) obtained from (a) ResNet-Duc, (b) SegNet, (c) ENet,  (d) U-Net, (e) FCN-8s, (f) cwGAN-gp, (g) PSPNet, (h) DeepLabv3+, (i) FC-DenseNet, (j) FC-DenseNet+FullCRF, (k) FC-DenseNet+FPCRF, and (l) ground truth from ISPRS dataset (spatial resolution: 5cm).}
    \label{Fig. 14}
  \end{figure*}

\begin{figure*}
     \subfloat[]{\includegraphics[width=0.22\textwidth]{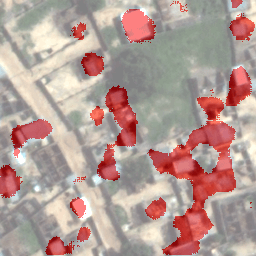}}
     \hspace*{\fill}
     \subfloat[]{\includegraphics[width=0.22\textwidth]{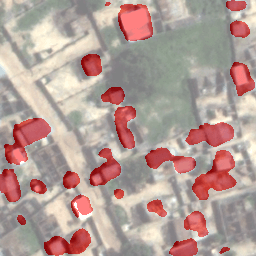}}
     \hspace*{\fill}
     \subfloat[]{\includegraphics[width=0.22\textwidth]{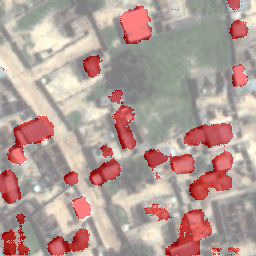}}
     \hspace*{\fill}
     \subfloat[]{\includegraphics[width=0.22\textwidth]{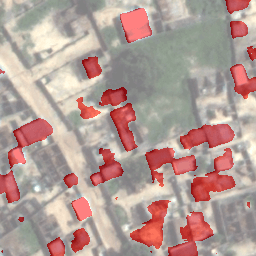}}
     \hfil
     \subfloat[]{\includegraphics[width=0.22\textwidth]{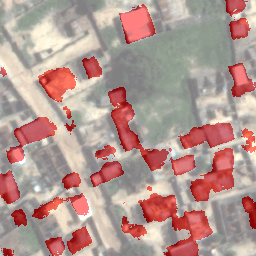}}
     \hspace*{\fill}
     \subfloat[]{\includegraphics[width=0.22\textwidth]{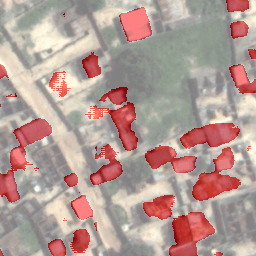}}
     \hspace*{\fill}
     \subfloat[]{\includegraphics[width=0.22\textwidth]{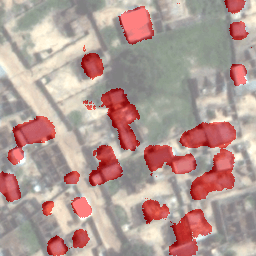}}
     \hspace*{\fill}
     \subfloat[]{\includegraphics[width=0.22\textwidth]{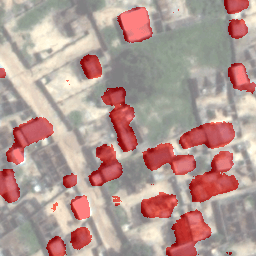}}
     \hfil
     \subfloat[]{\includegraphics[width=0.22\textwidth]{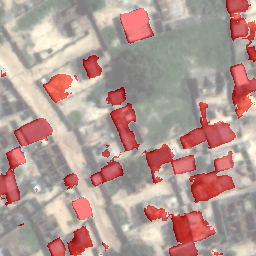}}
     \hspace*{\fill}
     \subfloat[]{\includegraphics[width=0.22\textwidth]{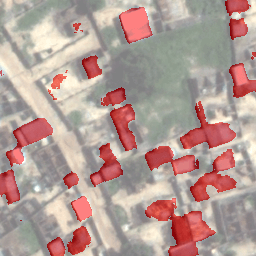}}
     \hspace*{\fill}
     \subfloat[]{\includegraphics[width=0.22\textwidth]{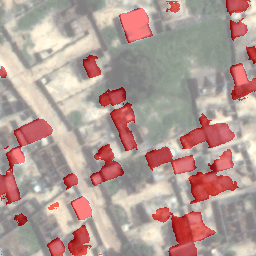}}
     \hspace*{\fill}
     \subfloat[]{\includegraphics[width=0.22\textwidth]{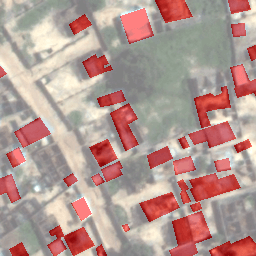}}
     \caption{The predicted results (in red) obtained from (a) ResNet-Duc, (b) SegNet, (c) ENet, (d) U-Net, (e) FCN-8s, (f) cwGAN-gp, (g) PSPNet, (h) DeepLabv3+, (i) FC-DenseNet, (j) FC-DenseNet+FullCRF, (k) FC-DenseNet+FPCRF, and (l) ground truth from Dstl dataset (spatial resolution: 1.24m).}
     \label{Fig. 15}
   \end{figure*}

\begin{figure*}
     \subfloat[]{\includegraphics[width=0.22\textwidth]{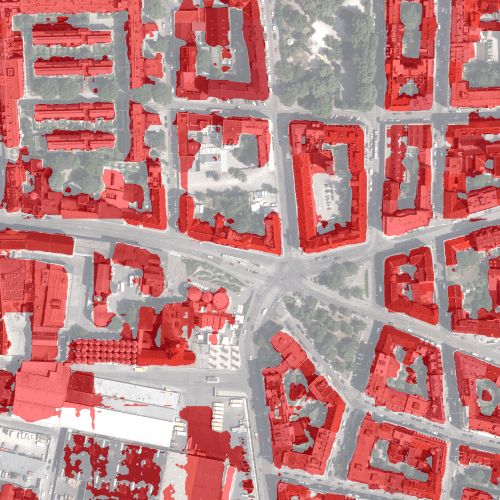}}
     \hspace*{\fill}
     \subfloat[]{\includegraphics[width=0.22\textwidth]{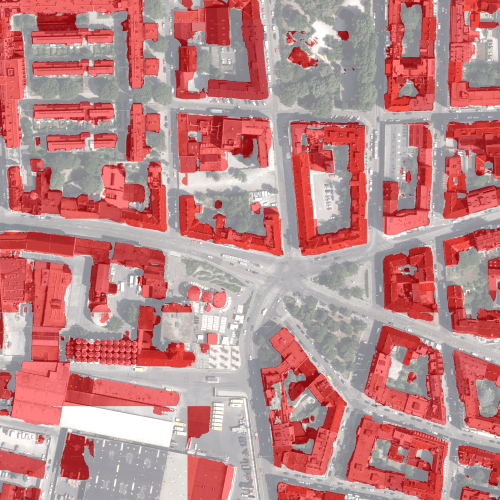}}
     \hspace*{\fill}
     \subfloat[]{\includegraphics[width=0.22\textwidth]{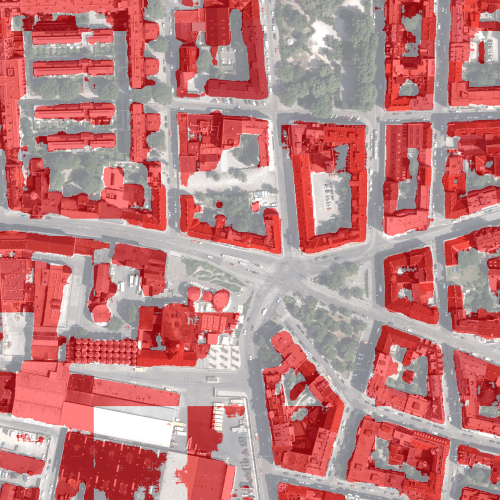}}
     \hspace*{\fill}
     \subfloat[]{\includegraphics[width=0.22\textwidth]{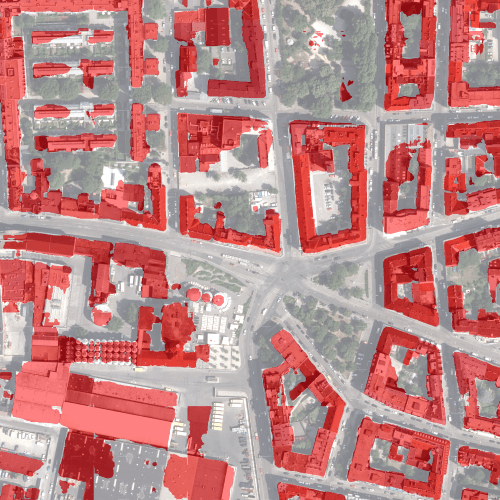}}
     \hfil
     \subfloat[]{\includegraphics[width=0.22\textwidth]{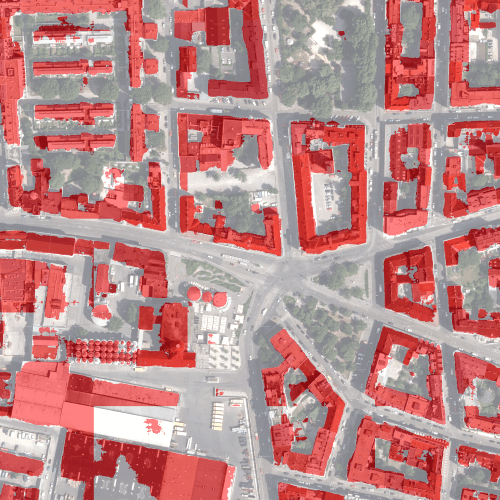}}
     \hspace*{\fill}
     \subfloat[]{\includegraphics[width=0.22\textwidth]{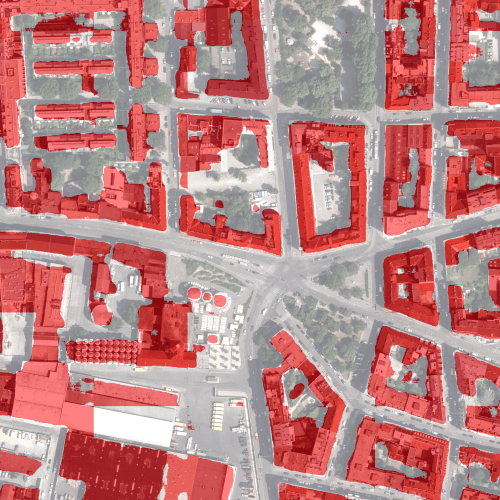}}
     \hspace*{\fill}
     \subfloat[]{\includegraphics[width=0.22\textwidth]{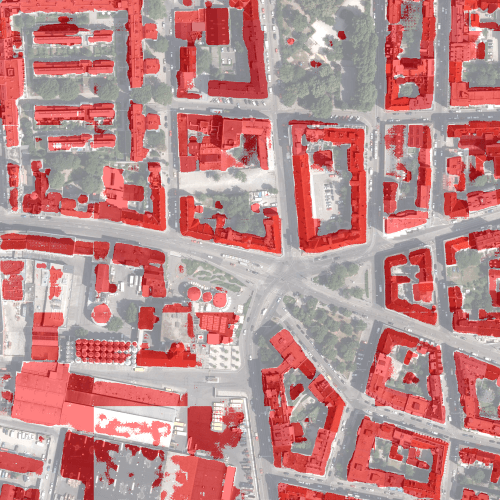}}
     \hspace*{\fill}
     \subfloat[]{\includegraphics[width=0.22\textwidth]{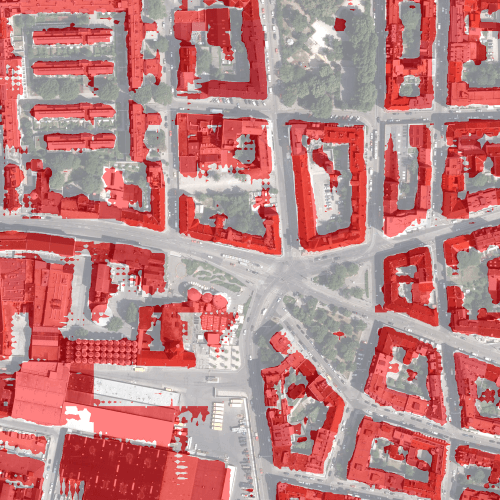}}
     \hfil
     \subfloat[]{\includegraphics[width=0.22\textwidth]{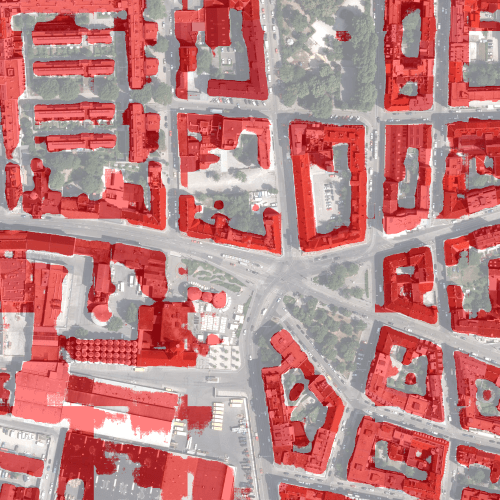}}
     \hspace*{\fill}
     \subfloat[]{\includegraphics[width=0.22\textwidth]{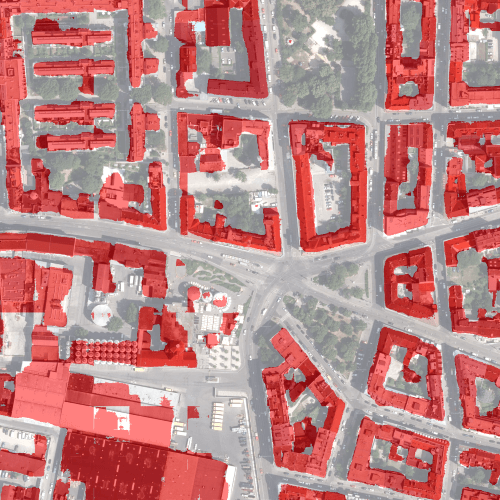}}
     \hspace*{\fill}
     \subfloat[]{\includegraphics[width=0.22\textwidth]{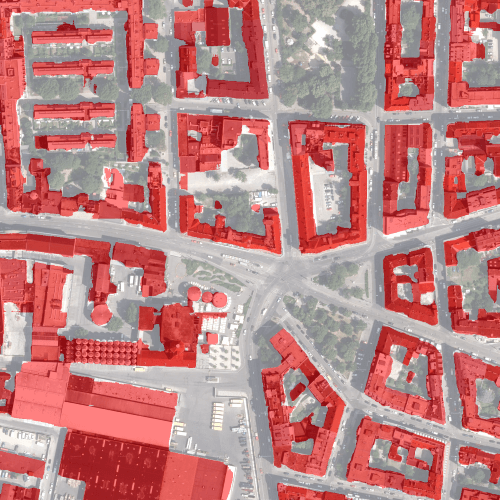}}
     \hspace*{\fill}
     \subfloat[]{\includegraphics[width=0.22\textwidth]{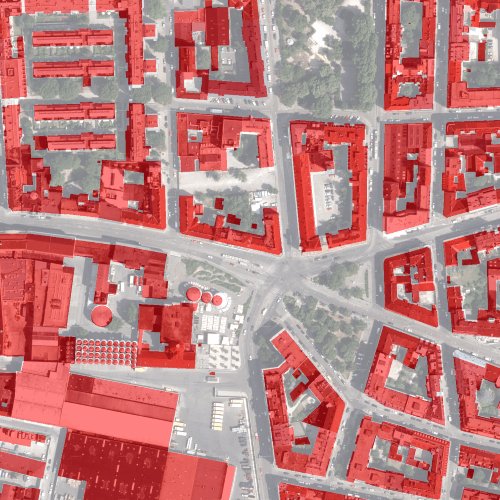}}
     \caption{The predicted results (in red) obtained from (a) ResNet-Duc, (b) SegNet, (c) ENet, (d) U-Net, (e) FCN-8s, (f) cwGAN-gp, (g) PSPNet, (h) DeepLabv3+, (i) FC-DenseNet, (j) FC-DenseNet+FullCRF, (k) FC-DenseNet+FPCRF, and (l) ground truth from Inria dataset (spatial resolution: 30cm).}
     \label{Fig. 16}
   \end{figure*}

\begin{figure*}[htbp]
 \captionsetup{justification=centering}
 \begin{center}
   \includegraphics[width=0.6\textwidth]{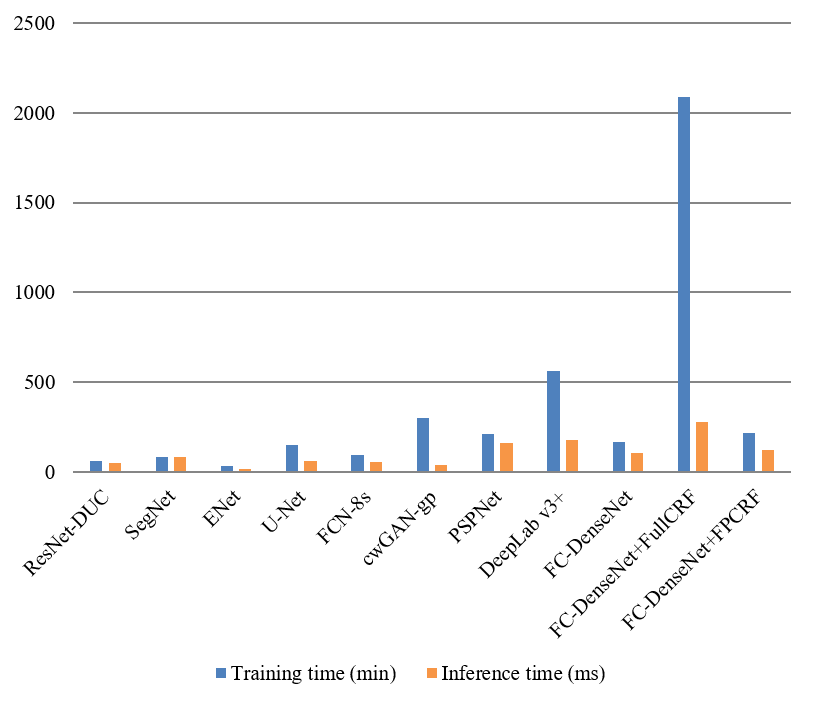}
 \end{center}
   \caption{Comparison of training time and inference time among different models from Planetscope dataset (spatial resolution: 3m)}
   \label{Fig. 17}
 \end{figure*}

In this research, several popular semantic segmentation neural networks from four different datasets were also investigated for comparisons of the proposed method. Their performance in building footprint generation, such as accuracy indexes are presented in Tables \ref{Tab. 5}, \ref{Tab. 6}, \ref{Tab. 7}, and \ref{Tab. 8}. Moreover, the visual results of different networks are also illustrated in Figs \ref{Fig. 13}, \ref{Fig. 14}, \ref{Fig. 15}, and \ref{Fig. 16}. The training and inference time costs of the different methods from Planetscope datatset are listed in the Fig. \ref{Fig. 17}, where the training time measures the whole training patches for 100 epochs, and inference time refers to the time cost for each patch.

DeepLabv3+ and PSPNet, which are the state-of-art networks for semantic segmentation tasks in computer vision, achieved satisfactory accuracy. These two networks are multiscale processing techniques, which not only allow the refinement of details, but also retain high-level semantic information. They can also take global structure into consideration when making local predictions. ENet is highly superior with respect to both training time and inference time, due to its specific architectures. First, the decoder uses max-pooling indices to produce sparse upsampled maps, which can reduce training time requirements. The input size can also be reduced heavily by the first two blocks, which adopt only a small number of features. Moreover, in the first stage, a max-pooling operation is performed in parallel with a strided convolution, and the resulting feature maps are concatenated, which speeds up inference process of the initial block. Compared to other CNN models, cwGAN-gp, which is a newly proposed network, also shows promising results for building footprint generation. The generator of cwGAN-gp exploits skip connection, which is helpful for retaining the boundary of the buildings. Moreover, the generator and discriminator of the GAN are both improved by the min-max game. However, the difficulty of training of GAN also leads to the longest training time among all the CNN models. Among all CNN models, FC-DenseNet is a superior network with respect to the numerical accuracy and visual results. On one hand, feature maps produced from different layers are concatenated in the DenseNet block, which can improve variation in the input of subsequent layers. On the other hand, high frequency information can be transferred by a standard skip connection between the encoder and the decoder, which contributes to the recovery of spatial details.

The architectures of the network, such as the feature extractor, decoder, and skip connection, have different significance when applied with satellite imagery of diverse spatial resolution. On one hand, for the higher spatial resolution satellite imagery (ISPRS dataset), the feature extractor is rather important. For instance, the accuracy indexes of PSPNet are much higher than those of DeepLabv3+, which means that the ResNet101 in PSPNet has a better feature extraction capability than the Xception in  DeepLabv3+. On the other hand, the decoder plays an important role in other datasets, including lower spatial resolution satellite imagery. DeepLabv3+ achieves much better results than PSPNet when applied in lower spatial resolution satellite imagery (Planetscope dataset, Dstl dataset, and Inria dataset), This is owing to the decoder module on top of the encoder output in DeepLabv3+, which contributes to sharper segmentation results. The skip connection in the networks (e.g., U-Net) is also vital to lower spatial resolution satellite imagery, as it is able to concatenate feature maps from both low-level and high-level layers. Hence, it can create a more efficient path for information propagation. However, it consumes more training and inference time, due to the fact that the feature maps from the encoder are transferred and concatenated to the decoder.

However, there are still some problems with CNN-based results such as weak boundaries and coarse pixel-level prediction. Therefore, graph models can be implemented to overcome the drawbacks of exploiting CNN for building footprint generation. CRF is a popular graph model with widespread success in solving semantic segmentation problems. The CRF inference can be used as a post-processing step, which is not integrated with the training of the CNN. However, in this case, the strength of CRF can not be fully harnessed. Therefore, we adopt an end-to-end deep learning network to produce sharp boundaries and fine-grained segmentation. FullCRF and FPCRF are combined with CNN models in one unified framework.  When connected with CRF-based graph models, the results can be improved as wrongly detected non-building pixels are removed. FC-DenseNet combined with FPCRF has achieved higher IoU and F1 scores than that combined with FullCRF, and can also better preserve the details and sharper boundaries. Moreover, FPCRF can substantially reduce the time needed for the training and inference stages. This superiority can be attributed to two reasons. First, FPCRF uses exact message passing, which avoids the approximation errors resulted from the permutohedral lattice approximation \cite{adams2010fast} in FullCRF. Second, localized processing in FPCRF can implement the feature learning more efficiently.

\begin{table}[htbp]
\captionsetup{justification=centering}
\caption{Comparison of accuracy indexes among different models of Planetscope dataset (spatial resolution: 3m)}
\begin{center}
\setlength\tabcolsep{2pt}
\begin{tabular}{llll}
  \hline\hline
   Models & Overall accuracy & F1 score & IoU\\
   \hline\hline
    ResNet-DUC & 0.7976 &	0.4593 & 0.2981\\
    SegNet & 0.8263	& 0.5597 & 0.3886\\
    ENet & 0.8379	& 0.5831 & 0.4115\\
    U-Net & 0.8435	& 0.6054	& 0.4341\\
    FCN-8s & 0.8505	& 0.6292	& 0.4590\\
    cwGAN-gp & 0.8453	& 0.6339	& 0.4641\\
    PSPNet & 0.8395 & 0.5948 &	0.4233 \\
    DeepLabv3+ & 0.8742 & 0.6592 &	0.4901\\
    FC-DenseNet & 0.8718	& 0.6556	& 0.4877\\
    FC-DenseNet+FullCRF & 0.8913	& 0.6580	& 0.4903\\
    \begin{bfseries}FC-DenseNet+FPCRF\end{bfseries} & \begin{bfseries}0.9297\end{bfseries} &  \begin{bfseries}0.6698\end{bfseries}	& \begin{bfseries}0.5046\end{bfseries}\\
     \hline
\end{tabular}
\end{center}
\label{Tab. 5}
\end{table}

 \begin{table}[htbp]
 \captionsetup{justification=centering}
 \caption{Comparison of accuracy indexes among different models of ISPRS dataset (spatial resolution: 5cm)}
 \begin{center}
 \setlength\tabcolsep{2pt}
 \begin{tabular}{llll}
   \hline\hline
    Models & Overall accuracy & F1 score & IoU\\
    \hline\hline
     ResNet-DUC & 0.7475 &	0.6766 & 0.5051\\
     SegNet & 0.8948 &	0.8511 & 0.7408\\
     ENet & 0.7711	& 0.7764 & 0.6110\\
     U-Net & 0.8892	& 0.8392 &	0.7229\\
     FCN-8s & 0.8617 &	0.7986 & 0.6647\\
     cwGAN-gp & 0.8926	& 0.8504	& 0.7397\\
     PSPNet & 0.9141 & 0.9144 &	0.8682\\
     DeepLabv3+ & 0.8995 &	0.9086 & 0.8325\\
     FC-DenseNet & 0.9186 &	0.9182 &	0.8789\\
     FC-DenseNet+FullCRF & 0.9298	& 0.9232	& 0.8826\\
     \begin{bfseries}FC-DenseNet+FPCRF\end{bfseries} & \begin{bfseries}0.9315\end{bfseries} &  \begin{bfseries}0.9358\end{bfseries}	& \begin{bfseries}0.8974\end{bfseries}\\
      \hline
 \end{tabular}
 \end{center}
 \label{Tab. 6}
 \end{table}

  \begin{table}[htbp]
  \captionsetup{justification=centering}
  \caption{Comparison of accuracy indexes among different models of Dstl dataset (spatial resolution: 1.24m)}
  \begin{center}
  \setlength\tabcolsep{2pt}
  \begin{tabular}{llll}
    \hline\hline
     Models & Overall accuracy & F1 score & IoU\\
     \hline\hline
      ResNet-DUC & 0.8923 &	0.5184 & 0.3499\\
      SegNet & 0.9240	& 0.6050 & 0.4337\\
      ENet & 0.9127	& 0.6890 & 0.5189\\
      U-Net & 0.9485	& 0.7576	& 0.5887\\
      FCN-8s & 0.9447	& 0.7467	& 0.5779\\
      cwGAN-gp & 0.9412	& 0.7291	& 0.5732\\
      PSPNet & 0.9379  & 0.6926   & 0.5297\\
      DeepLabv3+ & 0.9602 & 0.7578 & 0.6100\\
      FC-DenseNet & 0.9507	& 0.7602	& 0.5928\\
      FC-DenseNet+FullCRF & 0.9598	& 0.7697	& 0.6034\\
      \begin{bfseries}FC-DenseNet+FPCRF\end{bfseries} & \begin{bfseries}0.9604\end{bfseries} &  \begin{bfseries}0.7821\end{bfseries}	& \begin{bfseries}0.6176\end{bfseries}\\
       \hline
  \end{tabular}
  \end{center}
  \label{Tab. 7}
  \end{table}

\begin{table}[htbp]
\captionsetup{justification=centering}
\caption{Comparison of accuracy indexes among different models of INRIA dataset (spatial resolution: 30cm)}
\begin{center}
\setlength\tabcolsep{2pt}
\begin{tabular}{llll}
  \hline\hline
   Models & Overall accuracy & F1 score & IoU\\
   \hline\hline
    ResNet-DUC & 0.8704 &	0.7395 & 0.6097\\
    SegNet & 0.8826  & 0.7845 & 0.6455\\
    ENet & 0.8972 & 0.8001 & 0.6669\\
    U-Net & 0.9018  & 0.8027  & 0.6704\\
    FCN-8s & 0.9169 & 0.8192 & 0.6837\\
    cwGAN-gp & 0.9387	& 0.8371	& 0.7198\\
    PSPNet & 0.8960  & 0.7951  & 0.6599 \\
    DeepLabv3+ & 0.9498 & 0.8551 & 0.7299 \\
    FC-DenseNet & 0.9426 & 0.8536  & 0.7258\\
    FC-DenseNet+FullCRF & 0.9485	& 0.8605	& 0.7312\\
    \begin{bfseries}FC-DenseNet+FPCRF\end{bfseries} & \begin{bfseries}0.9581\end{bfseries} &  \begin{bfseries}0.8765\end{bfseries}	& \begin{bfseries}0.7479\end{bfseries}\\
     \hline
\end{tabular}
\end{center}
\label{Tab. 8}
\end{table}

\section{Conclusion}
Considering that there are weak boundary and coarse pixel-level label predictions in CNN-based results, we have proposed an end-to-end building footprint generation framework integrating CNN and a graph model in this research. Moreover, a number of the state-of-the-art CNN models for semantic segmentation are selected to generate building footprints from high resolution RS images for comparison. The effectiveness of  CNN models and the proposed end-to-end CNN-graph model building footprint generation approach are validated on four different datasets, (1) Planetscope satellite imagery of the cities of Munich, Paris, Rome, and Zurich; (2) aerial imagery of the City of Potsdam (North Germany) from ISPRS benchmark data; (3) WorldView3 satellite imagery from Dstl Kaggle dataset; (4) aerial imagery of the city of Austin, Chicago, Kitsap County, Western Tyrol, and Vienna from Inria Aerial Image Labeling data. The experimental results show that building footprint generation based on CNN-graph model-based methods can obtain more accurate results than CNN-based methods alone. Furthermore, FPCRF as the graph model in our proposed framework is effective in producing sharp boundaries and fine-grained segmentation results. On one hand, the completeness of the buildings can be preserved. On the other hand, some non-buildings, which are wrongly detected as buildings by CNN models, can be removed by graph models. Thus, we believe the proposed CNN-graph model method will be of practical value for the monitoring of fast-growing urban areas. In the future, we plan to extend our work to instance segmentation. More types of graph models will also be investigated.

\bibliography{Reference}
\bibliographystyle{IEEEtran}
\begin{IEEEbiography}
[{\includegraphics[width=1in,height=1.25in,clip,keepaspectratio]{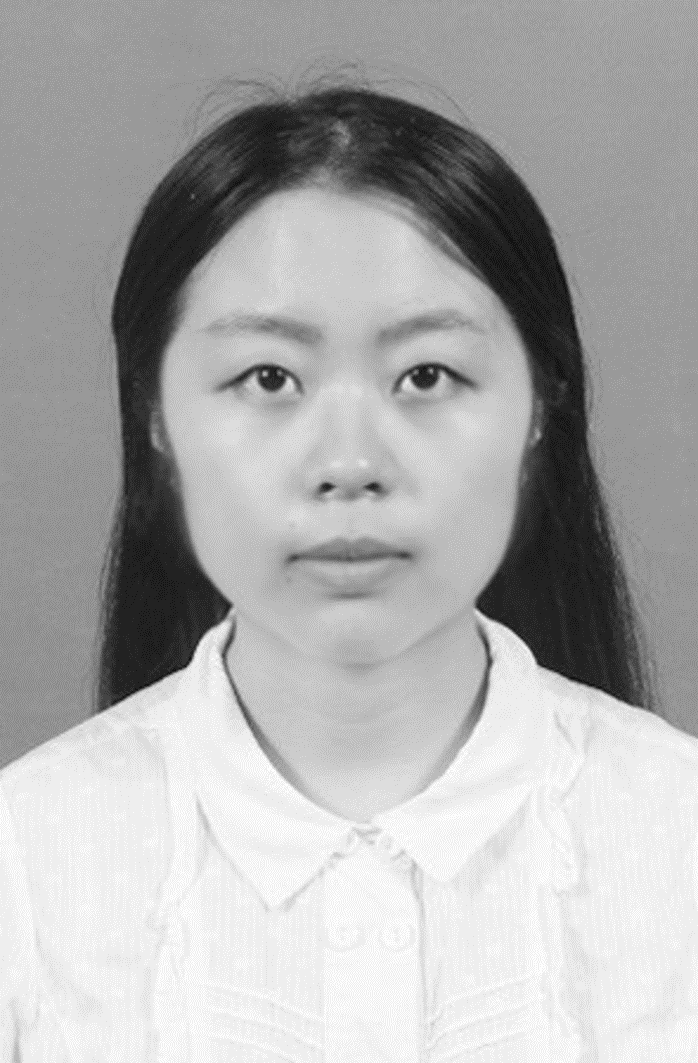}}] {Qingyu Li}
received the bachelor's degree in remote sensing science and technology from
the Wuhan University, Wuhan, China, in 2015, and
the master's degree in Earth Oriented Space Science and Technology (ESPACE) from the Technische Universit{\"a}t M{\"u}nchen (TUM), Munich, Germany, in
2018. 

She is currently pursuing the Ph.D. degree with
the German Aerospace Center (DLR), Wessling,
Germany and the TUM, Munich, Germany. Her research interests include deep learning, remote sensing mapping, and remote sensing applications.
\end{IEEEbiography}

\begin{IEEEbiography}
[{\includegraphics[width=1in,height=1.25in,clip,keepaspectratio]{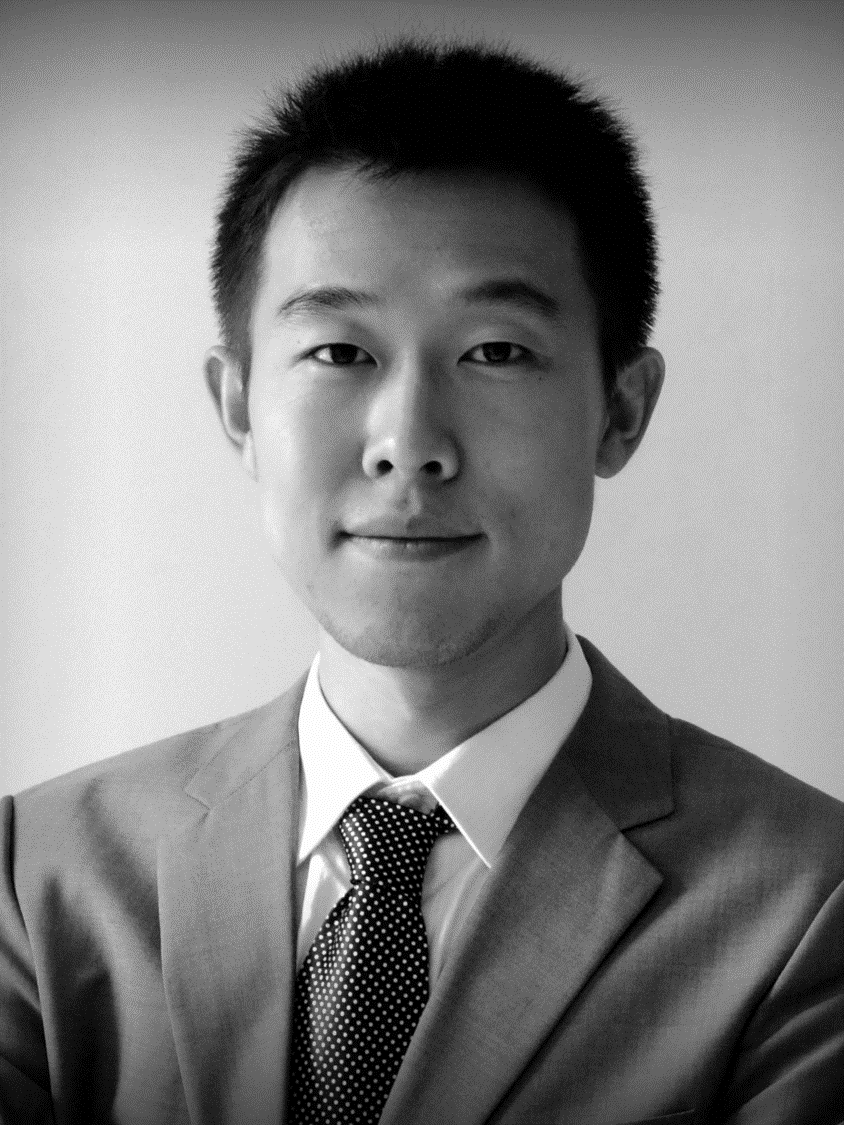}}] {Yilei Shi}
(M'18) received the Dipl.-Ing degree in
mechanical engineering and Dr.-Ing degree in signal processing from the Technische Universit{\"a}t M{\"u}nchen (TUM), Munich, Germany, in 2010 and 2019, respectively.

He is currently a senior scientist with the
Chair of Remote Sensing Technology, TUM. His
research interests include fast solver and parallel
computing for large-scale problems, high performance computing and computational intelligence, advanced
methods on SAR and InSAR processing, machine
learning and deep learning for variety of data
sources, such as SAR, optical images, and medical
images, and PDE-related numerical modeling and
computing.
\end{IEEEbiography}

\begin{IEEEbiography}[{\includegraphics[width=1in,height=1.25in,clip,keepaspectratio]{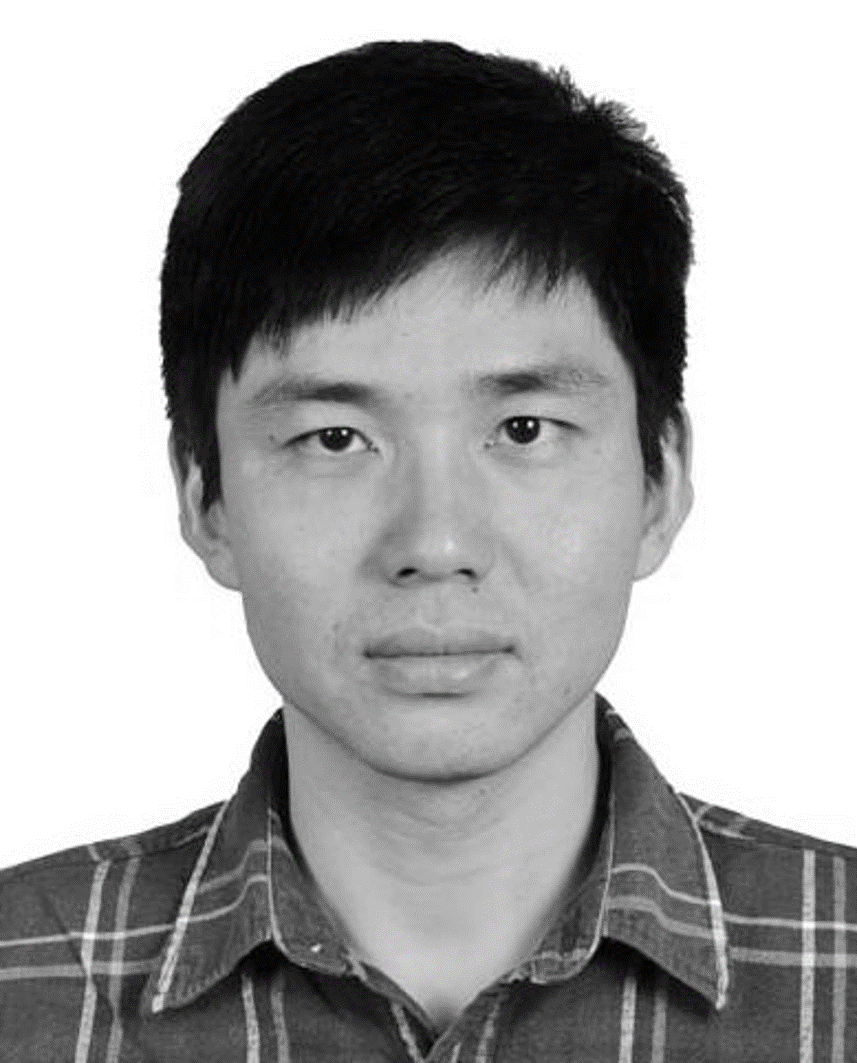}}] {Xin Huang} (M'13-SM'14)  received the Ph.D. degree in photogrammetry and remote sensing in 2009 from Wuhan University, Wuhan, China, working with the State Key Laboratory of Information Engineering in Surveying, Mapping and Remote Sensing (LIESMARS). 

He is currently a Luojia Distinguished Professor of Wuhan University, Wuhan, China, where he teaches remote sensing, photogrammetry, image interpretation, etc. He is the Founder and Director of the Institute of Remote Sensing Information Processing (IRSIP), School of Remote Sensing and Information Engineering, Wuhan University. He has published more than 140 peer-reviewed articles (SCI papers) in the international journals. His research interests include remote sensing image processing methods and applications. 
 
Prof. Huang has been supported by The National Program for Support of Top-notch Young Professionals (2017), the China National Science Fund for Excellent Young Scholars (2015), and the New Century Excellent Talents in University from the Ministry of Education of China (2011). He was the recipient of the Boeing Award for the Best Paper in Image Analysis and Interpretation from the American Society for Photogrammetry and Remote Sensing (ASPRS) in 2010, the second place recipient for the John I. Davidson President’s Award from ASPRS in 2018, and the National Excellent Doctoral Dissertation Award of China in 2012. In 2011, he was recognized by the IEEE Geoscience and Remote Sensing Society (GRSS) as the Best Reviewer of IEEE GEOSCIENCE AND REMOTE SENSING LETTERS. He was the winner of the IEEE GRSS 2014 Data Fusion Contest. Prof. Huang was the lead guest editor of the special issue for the IEEE JOURNAL OF SELECTED TOPICS IN APPLIED EARTH OBSERVATIONS AND REMOTE SENSING (May 2015 and Aug 2019), the Journal of Applied Remote Sensing (Oct 2016), Photogrammetric Engineering and Remote Sensing (Nov 2018), and Remote Sensing (Nov 2019). He was an Associate Editor of the Photogrammetric Engineering and Remote Sensing (2016-2019), and now serves as an Associate Editor of the IEEE GEOSCIENCE AND REMOTE SENSING LETTERS (since 2014), and an Associate Editor of the IEEE JOURNAL OF SELECTED TOPICS IN APPLIED EARTH OBSERVATIONS AND REMOTE SENSING (since 2018). He is also an editorial board member of the Remote Sensing of Environment (since 2019), and the Remote Sensing (An open access journal from MDPI) since 2018.

\end{IEEEbiography}

\begin{IEEEbiography}[{\includegraphics[width=1in,height=1.25in,clip,keepaspectratio]{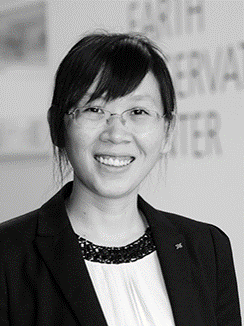}}]{Xiao Xiang Zhu}
(S'10--M'12--SM'14) received the Master (M.Sc.) degree, her doctor of engineering (Dr.-Ing.) degree and her “Habilitation” in the field of signal processing from Technical University of Munich (TUM), Munich, Germany, in 2008, 2011 and 2013, respectively.
\par
  She is currently the Professor for Signal Processing in Earth Observation (www.sipeo.bgu.tum.de) at Technical University of Munich (TUM) and the Head of the Department ``EO Data Science'' at the Remote Sensing Technology Institute, German Aerospace Center (DLR). Since 2019, Zhu is co-coordinating the Munich Data Science Research School (www.mu-ds.de). She is also leading the Helmholtz Artificial Intelligence Cooperation Unit (HAICU) -- Research Field ``Aeronautics, Space and Transport". Prof. Zhu was a guest scientist or visiting professor at the Italian National Research Council (CNR-IREA), Naples, Italy, Fudan University, Shanghai, China, the University  of Tokyo, Tokyo, Japan and University of California, Los Angeles, United States in 2009, 2014, 2015 and 2016, respectively. Her main research interests are
  remote sensing and Earth observation, signal processing, machine learning and data science, with a special application focus on global urban mapping.

  Dr. Zhu is a member of young academy (Junge Akademie/Junges Kolleg) at the Berlin-Brandenburg Academy of Sciences and Humanities and the German National  Academy of Sciences Leopoldina and the Bavarian Academy of Sciences and Humanities. She is an associate Editor of IEEE Transactions on Geoscience and Remote Sensing.
\end{IEEEbiography}
\end{document}